\DocumentMetadata{}
\documentclass[sigconf]{acmart}
\pdfoutput=1
\copyrightyear{2026}
\acmYear{2026}
\setcopyright{cc}
\setcctype{by}
\acmConference[KDD '26]{Proceedings of the 32nd ACM SIGKDD Conference on Knowledge Discovery and Data Mining V.1}{August 09--13, 2026}{Jeju Island, Republic of Korea}
\acmBooktitle{Proceedings of the 32nd ACM SIGKDD Conference on Knowledge Discovery and Data Mining V.1 (KDD '26), August 09--13, 2026, Jeju Island, Republic of Korea}
\acmPrice{}
\acmDOI{10.1145/3770854.3780312}
\acmISBN{979-8-4007-2258-5/2026/08}

\usepackage{hyperref}
\hypersetup{colorlinks=true, linkcolor=blue, anchorcolor=blue, citecolor=blue}
\usepackage{amsmath}
\usepackage{graphicx}
\usepackage{caption}
\usepackage{subcaption}
\usepackage{colortbl}
\usepackage[export]{adjustbox} 
\usepackage{bbding}
\usepackage{multirow}
\usepackage{pifont}

\usepackage{bm}

\usepackage{enumitem}
\usepackage[ruled,vlined,linesnumbered]{algorithm2e}
\usepackage{xspace}
\usepackage{makecell}
\usepackage{lipsum}
\usepackage{balance} 
\usepackage{adjustbox}

\newcommand{\tableref}[1]{Table~\ref{#1}} 

\newcommand{\figref}[1]{Figure~\ref{#1}} 

\AtBeginDocument{%
	\providecommand\BibTeX{{%
			\normalfont B\kern-0.5em{\scshape i\kern-0.25em b}\kern-0.8em\TeX}}}

\begin{document}
	\newcommand{\ndatasets}{{25}\xspace}
	\title{Towards Resilient Transportation: A Conditional Transformer for Accident-Informed Traffic Forecasting}
	
	\author{Hongjun Wang}
	\affiliation{%
		\institution{The University of Tokyo}
		\city{Tokyo}
		\country{Japan}
	}
	
	\author{Jiawei Yong}
	\affiliation{%
		\institution{Toyota Motor Corporation}
		\city{Tokyo}
		\country{Japan}
	}
	
	\author{Jiawei Wang}
	\affiliation{%
		\institution{The University of Tokyo}
		\city{Tokyo}
		\country{Japan}
	}
	
	\author{Shintaro Fukushima}
	\affiliation{%
		\institution{Toyota Motor Corporation}
		\city{Tokyo}
		\country{Japan}
	}
	
	\author{Renhe Jiang}
	\authornote{Corresponding author.}
	\email{jiangrh@csis.u-tokyo.ac.jp}
	\affiliation{%
		\institution{The University of Tokyo}
		\city{Tokyo}
		\country{Japan}
	}
	
	\renewcommand{\shortauthors}{Hongjun Wang et al.}

	\begin{abstract}
		Traffic prediction remains a key challenge in spatio-temporal data mining, despite progress in deep learning. Accurate forecasting is hindered by the complex influence of external factors such as traffic accidents and regulations, often overlooked by existing models due to limited data integration. 
		To address these limitations, we present two enriched traffic datasets from Tokyo and California, incorporating traffic accident and regulation data. Leveraging these datasets, we propose ConFormer (Conditional Transformer), a novel framework that integrates graph propagation with guided normalization layer. This design dynamically adjusts spatial and temporal node relationships based on historical patterns, enhancing predictive accuracy. 
		Our model surpasses the state-of-the-art STAEFormer in both predictive performance and efficiency, achieving lower computational costs and reduced parameter demands. Extensive evaluations demonstrate that ConFormer consistently outperforms mainstream spatio-temporal baselines across multiple metrics, underscoring its potential to advance traffic prediction research.\textcolor{magenta}{\textit{The code is released in  \url{https://github.com/Dreamzz5/ConFormer}}}. 
		\end{abstract}



\maketitle

\section{Introduction}
\begin{figure}
\centering
\subfloat[Traffic states under normal and abnormal conditions.]{\includegraphics[width=1\linewidth]{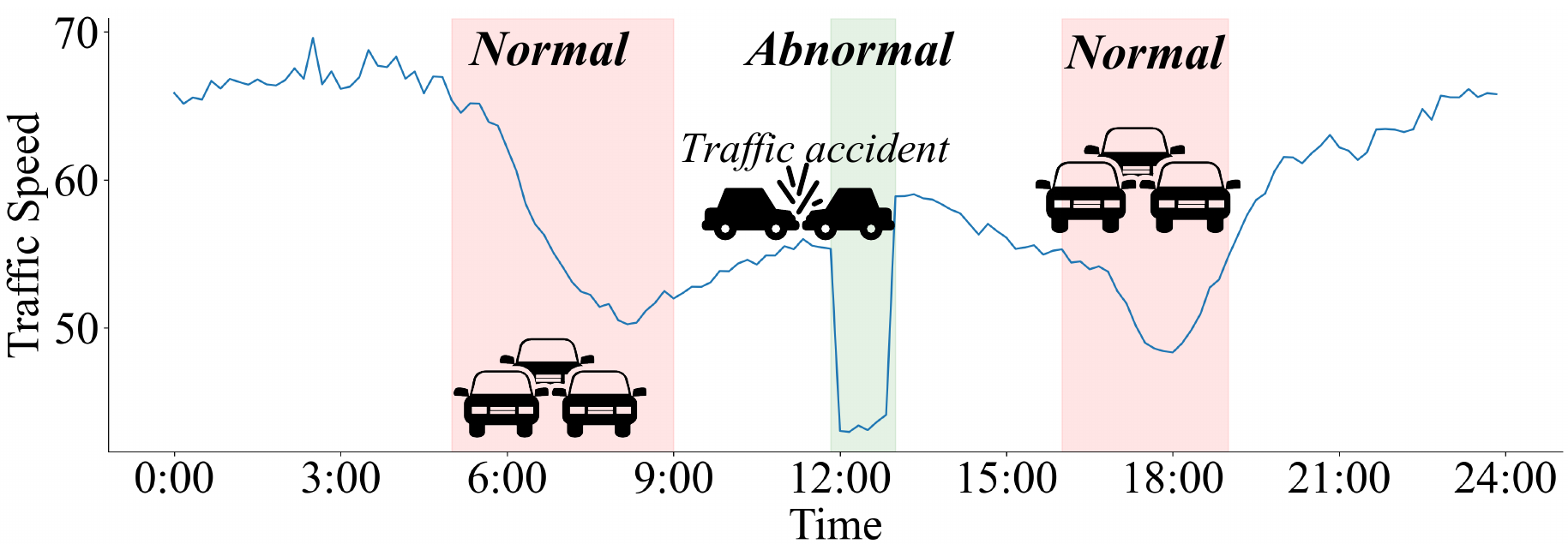} \label{fig:1a}}\\
\subfloat[Accidents by severity level in the California and Tokyo regions.]{\includegraphics[width=\linewidth]{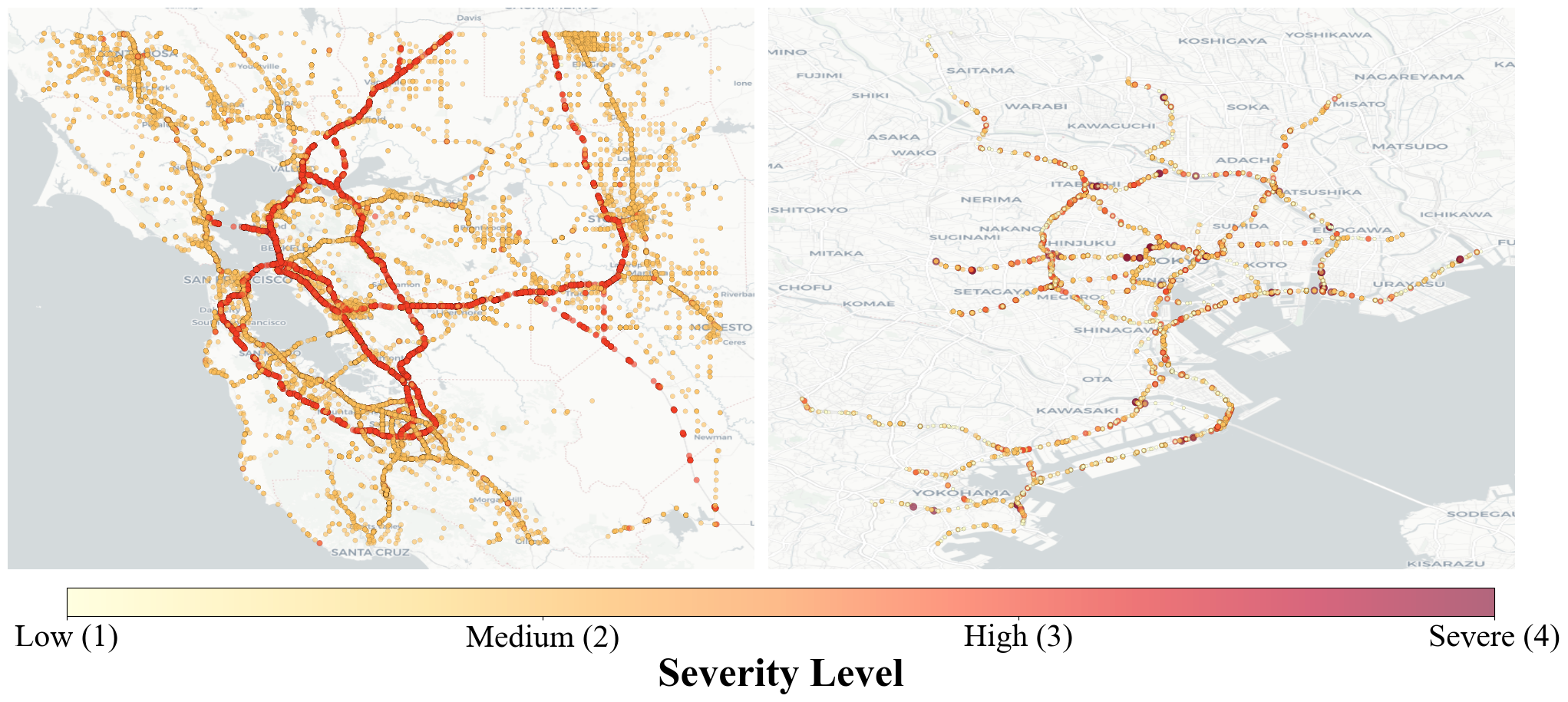} \label{fig:1b}}
\caption{Illustration of our motivation: enhancing traffic forecasting towards resilient transportation.}
\vspace{-10pt}
\label{fig:bar}
\end{figure}

Traffic forecasting remains a key challenge in intelligent transportation systems, supporting route planning, emergency response, and urban traffic management. While deep learning has made major progress in recent years, predicting traffic conditions still faces a critical problem: existing methods work well for regular patterns but struggle when traffic accidents disrupt normal flow. As shown in Figure~\ref{fig:1a}, traffic systems follow predictable patterns during normal conditions (like rush hours), but traffic accidents cause sudden speed drops and complex changes that current spatiotemporal methods cannot handle well. Studies show that traffic incidents greatly reduce prediction accuracy, with highway accidents contributing significantly to overall travel delays~\cite{habtemichael2015incident, chung2011quantification}. Real-world data shows that accidents can increase travel times by 37-43\% compared to normal conditions~\cite{fhwa2005congestion, ops2004linking}.

Traffic systems show a basic conflict between predictable and unpredictable behavior. Normal traffic follows known time patterns, with rush hour congestion well captured by models like DCRNN~\cite{DCRNN}. However, accidents create complex disturbances that spread through connected road networks in nonlinear ways~\cite{GWNet, zhang2020curb}, producing severe disruptions beyond what current spatiotemporal methods can model. This reveals two main problems in current approaches: (1) \textbf{Limited accident data coverage}. Most traffic datasets lack detailed incident information, offering only basic flow data, which prevents systematic study of how accidents and regulations affect traffic patterns~\cite{ben2010theory}; (2) \textbf{Inadequacies in accident modeling}. Current approaches ignore incident factors entirely, failing to account for how accidents change spatial relationships, time dependencies, and data patterns during disrupted conditions~\cite{GWNet, STID}.

To address these fundamental limitations, first, we create two comprehensive datasets integrating detailed incident information with traditional flow measurements. Our Tokyo dataset covers 1,843 highway segments with accident and regulatory data from JARTIC \cite{jartic2025}, while our California dataset encompasses Bay Area (2,352 sensors) and San Diego (716 sensors) with comprehensive accident classifications from the US Accidents database \cite{moosavi2019countrywide}. As demonstrated in \figref{fig:1b}, accident distributions exhibit significant geographic variation in density and severity patterns across the Bay Area and Tokyo regions. 

Building upon recent advances in conditional modeling \cite{rombach2022high, yang2023diffusion}, we propose ConFormer, a novel conditional Transformer architecture for incident-aware traffic prediction. The core innovation lies in our Guided Layer Normalization (GLN) mechanism, which replaces static normalization parameters with dynamic affine transformations conditioned on prevailing traffic conditions \cite{park2019semantic}. This enables adaptive modulation of internal feature distributions during incident scenarios where conventional normalization proves inadequate. ConFormer integrates GLN with graph propagation mechanisms that diffuse incident information across traffic networks, capturing complex spatiotemporal correlations when localized incidents propagate through interconnected road segments \cite{kipf2017semi}. Despite sophisticated conditioning mechanisms, ConFormer maintains computational efficiency with reduced parameters and overhead compared to STAEFormer~\cite{STAEformer}, achieving superior predictive accuracy while preserving scalability for real-world deployment.
We make the following contributions:
\begin{itemize}[leftmargin=0.4cm]
    \item \textbf{ConFormer}, a conditional Transformer employing guided layer normalization to adaptively capture spatiotemporal dynamics with incident awareness;
    \item \textbf{Two large-scale datasets}: Tokyo highways with accident and regulatory data, and California highways with comprehensive accident annotations;
    \item \textbf{State-of-the-art performance} surpassing STAEFormer across normal and incident scenarios while maintaining superior computational efficiency with fewer parameters and lower overhead.
\end{itemize}


\section{Related Works}
Spatiotemporal forecasting has garnered significant attention due to its critical importance in various real-world applications. Traditional approaches such as ARIMA~\cite{ARIMA-traffic}, VAR~\cite{VAR}, $k$-NN~\cite{kNN-traffic}, and SVM~\cite{SVM-traffic} often struggle to effectively capture the complex dependencies inherent in spatiotemporal data.  
Deep learning has led to substantial advancements in spatiotemporal forecasting. Recurrent Neural Networks, including LSTMs~\cite{LSTM-traffic-2015, LSTM-traffic-2018} and GRUs~\cite{GRU}, have shown considerable success in modeling temporal dynamics. However, these models frequently fall short in capturing spatial dependencies, which are essential for accurate predictions in networked urban systems.
To address this shortcoming, the integration of Graph Convolutional Networks (GCNs) with temporal models has been proposed, resulting in notable improvements in spatiotemporal forecasting. Prominent examples include STGCN~\cite{stgcn} and DCRNN~\cite{DCRNN}, which have laid the groundwork for further advancements in the field~\cite{GWNet, AGCRN, StemGNN, MTGNN}. Additionally, a variety of innovative methods have been developed~\cite{GTS, STGNN, PMMemNet, MegaCRN, CaST}, including meta-parameter learning approaches~\cite{dong2024heterogeneity}, normalization techniques~\cite{deng2021stnorm}, large-scale forecasting methods~\cite{han2024bigst}, expert frameworks~\cite{wang2022st}, and curriculum learning strategies~\cite{wang2023easy}.
Time series transformers~\cite{TFTimeSeriesSurvey, chen2024multi} have also been employed to capture spatiotemporal correlations~\cite{GMAN, PDFormer, STAEformer, TESTAM} and manage long sequences~\cite{Informer, Autoformer, FEDformer, iTransformer}. Recent advances include masked pre-training strategies for spatiotemporal forecasting~\cite{gao2024spatial} and efficient transformer architectures~\cite{wang2024stgformer}. Although traffic
accidents significantly influence traffic propagation, the extent and
mechanisms of these effects remain insufficiently explored. While some studies have addressed multi-incident co-prediction~\cite{wang2021spatio} and traffic accident risk prediction~\cite{zhang2025uncertainty, zhou2022foresee}, existing approaches often treat accidents as external factors rather than integrated components of the forecasting model. \textit{To address
this gap, our study proposes a novel conditional transformer
architecture, which integrates guided normalization layers to dynamically
adjust node spatial relationships and temporal correlations in
response to traffic accidents, providing a more resilient and powerful
traffic forecasting technique.}

\section{PROBLEM STATEMENT}
The traffic network is modeled as a graph $G = (\mathcal{V}, \mathcal{E}, \mathcal{A})$, where $\mathcal{V}$ represents $N$ nodes, $\mathcal{E}$ denotes edges, and $\mathcal{A} \in \mathbb{R}^{N \times N}$ encodes node relationships. Traffic states over past $T$ time steps observed at time $t$ are expressed as $X_t \in \mathbb{R}^{T \times N \times D}$, where $D$ is the feature dimension (e.g., flow or speed). In this study, dynamic traffic conditions such as traffic accident or regulation are incorporated as guided features. Our target is to predict the traffic state for the future $T'$ time steps $\hat{Y}_t \in \mathbb{R}^{T' \times N \times 1}$, where $\hat{Y}_t$ denotes the predicted traffic states, thus our problem is defined as $[X_t; \mathcal{A}] \xrightarrow{q(\cdot)} (\beta, \gamma, \alpha)$ and $[X_t; \beta, \gamma, \alpha] \xrightarrow{f(\cdot)} \hat{Y}_t$, where $q(\cdot)$ computes global factors $\beta, \gamma, \alpha$ to guide the prediction model $f(\cdot)$.


\section{Methodology}
\begin{figure}[t]
\centering
\includegraphics[width=1\linewidth]{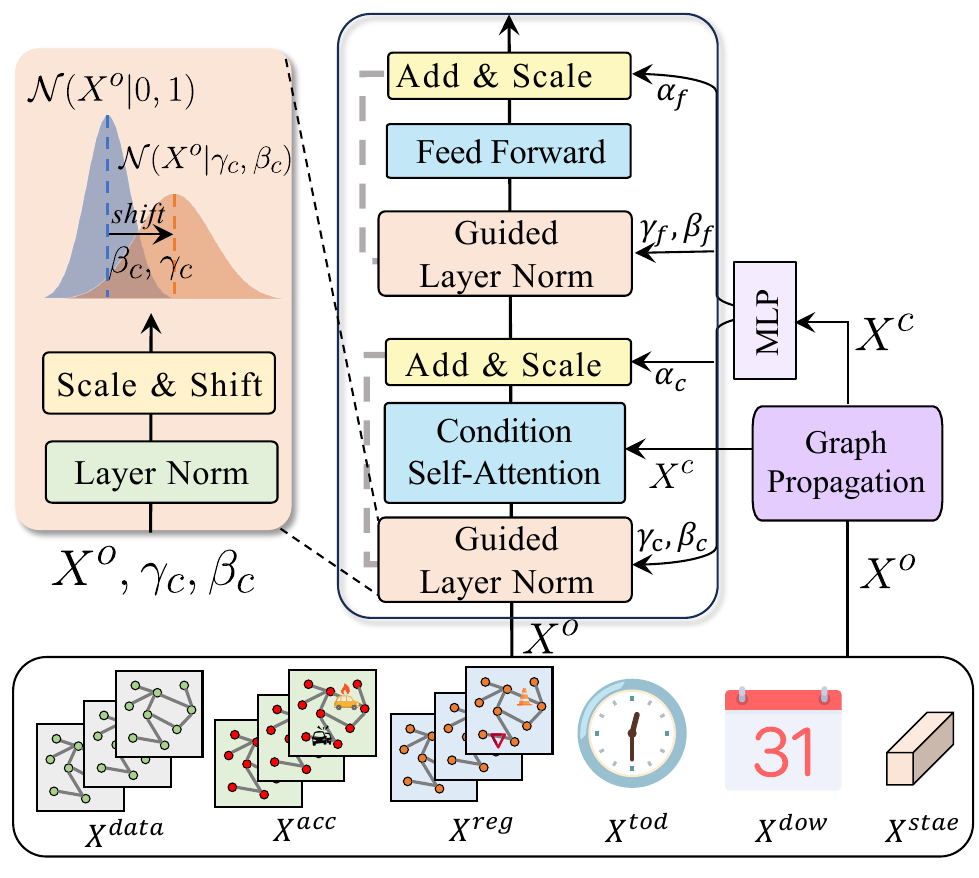}
\caption{ConFormer enhances the spatiotemporal transformer with Graph Propagation and Guided LayerNorm (GLN). The graph propagation, combined with an MLP, generates the mean (\(\beta\)), variance (\(\gamma\)), and amplitude factor (\(\alpha\)) to guide layer normalization and residual connections, regulating the latent space distribution at each node.}
\label{fig:framework}
\end{figure}

The overall architecture of our proposed ConFormer is illustrated as Figure \ref{fig:framework}. We expand its details in the following sections.
\subsection{Spatiotemporal Condition Propagation}\label{sec:trans}
The data embedding layer transforms input observations $\bm{X}_t$ into a high-dimensional representation ${X}^{data} \in \bm{R}^{T \times N \times D_{data}}$ via a fully connected layer, where $D_{data}$ is the embedding dimension. To incorporate domain knowledge, we design spatiotemporal embeddings that capture both network structure and traffic periodicity.
Following the concatenation order in our final embedding, we first incorporate traffic event indicators $X^{acc} \in \bm{R}^{T \times N \times D_{acc}}, X^{reg} \in \bm{R}^{T \times N \times D_{reg}}$ for accidents and regulations respectively. To model traffic periodicity patterns, we introduce weekly and daily embeddings $\omega(t) \rightarrow {X}^{dow} \in \bm{R}^{T \times N \times D_{dow}}, \tau(t) \rightarrow {X}^{tod} \in \bm{R}^{T \times N \times D_{tod}}$, where $\omega(t)$ maps to week indices (1-7) and $\tau(t)$ to time-interval indices (e.g., 1-288 for 5-minute intervals in a day). The temporal embeddings $X^{dow}, X^{tod}$ are constructed by concatenating embeddings across all time steps $T$. Moreover, we incorporate spatiotemporal adaptive embedding $X^{stae} \in \bm{R}^{T \times N \times D_{stae}}$ ~\cite{STAEformer} and concatenate all the embeddings together as follows:
\begin{equation}
\bm{X}^{o} = \texttt{MLP}(X^{data} \ || \ X^{acc} \ || \ X^{reg} || \ X^{dow} \ || \ X^{tod} \ || \ X^{stae} ),
\end{equation}
where \(\parallel\) denotes the concatenation operation.

\noindent\textbf{Graph Propagation.} To model the accident propagation effect, we leverage the idea of graph propagation from GCN \cite{kipf2017semi}, which operate over a graph $\mathcal{G}$. Given a graph signal $X \in \bm{R}^{|\mathcal{V}| \times D}$, where $D$ is the number of node features, a typical $K$-hop graph convolutional layer is formulated as $G C N_{\mathcal{G}}(X ; W, \theta)=\sigma\left(\sum_{k=0}^K \theta_k \mathcal{L}^k X\right) W$, where $\mathcal{L} \in[0,1]^{|\mathcal{V}| \times|\mathcal{V}|}$ is the graph Laplacian of the adjacency matrix $\mathcal{A} $that controls information propagation between nodes; $\theta \in \bm{R}^K$  weights contributions from different hops; $W \in \bm{R}^{N \times D'}$ ($D'$ is the output feature dimension) mixes feature dimensions; and $\sigma$ denotes the activation function. In our work, to model the condition propagation effect (i.e., the accident impact), we simplify the GCN by applying only the graph propagation and omitting the feature mixing with $W$ on the input embedding $\bm{X}^{o}$, as follows:
\begin{equation}\label{eq:graphpropagation}
    \begin{aligned}
\bm{X}^c &= \texttt{GraphPropagation}(\bm{X}^o, \mathcal{A}) \\ 
         &= \left[\bm{X}^{o} \ \| \ \mathcal{L} \bm{X}^{o}\| \ \mathcal{L}^2 \bm{X}^{o} \ \| \ \ldots \ \| \ \mathcal{L}^K \bm{X}^{o}\right].
\end{aligned}
\end{equation}
$\bm{X}^c$ can be regarded as a contextual condition representation that incorporates the propagation effect.

\subsection{Spatiotemporal Conditional Transformer}\label{sec:cond}
The spatiotemporal Transformer architecture has shown superior performance in traffic forecasting~\cite{STAEformer}; therefore, we adopt it as the backbone of our model. Given the traffic input embedding $\bm{X}^{o}$,  the query, key, and value for the self-attention mechanism are computed as follows:
\begin{equation}\label{eq:self-attention}
\bm{Q} = \bm{X}^o \bm{W}_Q, \quad
\bm{K} = (\bm{X}^o ) \bm{W}_K, \quad
\bm{V} = (\bm{X}^o ) \bm{W}_V,
\end{equation}
where \(\bm{W}_Q, \bm{W}_K, \bm{W}_V\) are learnable parameters. Then, for spatial dependencies, at each time step $t$, we compute the spatial attention scores as: $ \bm{A}^{(Sp)}_t = \frac{\bm{Q}_{t,:,:}(\bm{K}_{t,:,:})^{\top}}{\sqrt{d'}} $ where $\bm{Q}_{t,:,:}, \bm{K}_{t,:,:}, \bm{V}_{t,:,:} \in \bm{R}^{N \times d'}$ are the query, key, and value matrices for the spatial dimension. The output of the spatial self-attention is then obtained as: $ \bm{X}^{(Sp)} = \mathrm{softmax}(\bm{A}^{(Sp)}_t)\bm{V}_{t,:,:} $ Similarly, for temporal dependencies, we compute the temporal attention scores for each node $n$ as: $ \bm{A}^{(Te)}_n = \frac{\bm{Q}_{:,n,:}(\bm{K}_{:,n,:})^{\top}}{\sqrt{d'}} $ where $\bm{Q}_{:,n,:}, \bm{K}_{:,n,:}, \bm{V}_{:,n,:} \in \bm{R}^{T \times d'}$ are the query, key, and value matrices for the temporal dimension. The output of the temporal self-attention is then obtained as: $ \bm{X}^{(Te)} = \mathrm{softmax}(\bm{A}^{(Te)}_n)\bm{V}_{:,n,:}$. Finally, an MLP is employed to fuse $\bm{X}^{(Sp)}$ and $\bm{X}^{(Te)}$ into a unified spatiotemporal representation:
\begin{align}\label{eq:st-attention}
\bm{X}^{(Att)} = \texttt{MLP}\left(\bm{X}^{(Sp)} \parallel \bm{X}^{(Te)}\right).
\end{align}
Although the backbone is capable of capturing spatiotemporal dependencies, real-world traffic states are inherently dynamic and complex. Factors such as traffic accidents, local regulations, and sudden demand fluctuations introduce substantial variations that are challenging to model. To address this challenge, we propose \textit{Guided Layer Normalization (GLN)} to substitute the classic $\text{LayerNorm}$~\cite{ba2016layer}, which adapts the normalization process using dynamic, condition-dependent parameters derived from the contextual condition representation.


\noindent\textbf{Guided Layer Normalization.}
$\text{LayerNorm}$~\cite{ba2016layer} is widely used to stabilize the training of deep neural networks by normalizing inputs across features. Given an input $\bm{X}^o$, it is computed as: \(\bm{X}^o = \gamma \frac{\bm{X}^o - \mu}{\sigma} + \beta\), where $\mu$ and $\sigma$ are the mean and standard deviation of the elements in $\bm{X}^o$, and $\gamma$ and $\beta$ are learnable but fixed parameters shared across all inputs. Instead, we propose to learn dynamic affine transformation factors $\gamma, \beta$ from the condition representation $\bm{X}^c$ and guide the normalization as follows:
\begin{equation}
\begin{aligned}\label{eq:GLN}
 \gamma, \beta &= \texttt{MLP}(\bm{X}^c), \\
 \texttt{GLN}(\bm{X}^o, \gamma, \beta) &=\gamma \left(\frac{\bm{X}^o-\mu}{\sigma}\right)+ \beta. 
\end{aligned}
\end{equation}
With GLN, the vanilla self-attention (applied along spatial/temporal axis)  can be reformulated as follows:
\begin{equation}
	\adjustbox{max width=\linewidth}{
		$\operatorname{Softmax}\left( \frac{\mathbf{Q} \mathbf{K}^\top}{\sqrt{D_k}} \right) \rightarrow \operatorname{Softmax}\left( \frac{\gamma^2 \mathbf{Q}' \mathbf{K}'^\top + \gamma \mathbf{K}' \beta^\top + \gamma \mathbf{Q}' \beta^\top + \beta^2}{\sqrt{D_k}} \right)$
	}
\end{equation}
where \(\mathbf{Q}' = \gamma \cdot \frac{\mathbf{Q} - \boldsymbol{\mu}_{\mathbf{Q}}}{\boldsymbol{\sigma}} + \beta\) and \(\mathbf{K}' = \gamma \cdot \frac{\mathbf{K} - \boldsymbol{\mu}_{\mathbf{K}}}{\boldsymbol{\sigma}} + \beta\). The proof of this reformulation is provided in Appendix. Based on this formulation, we can derive the following interpretations:
\begin{itemize}[leftmargin=0.4cm]
    \item \textbf{For $\gamma$ (scaling factor):}
    \begin{itemize}
        \item \textbf{Higher $\gamma$:} Increases sensitivity to abrupt changes in traffic patterns, enabling rapid adaptation to accidents or anomalies.
        \item \textbf{Lower $\gamma$:} Suppresses the influence of local fluctuations, promoting stability under regular traffic conditions.
    \end{itemize}

    \item \textbf{For $\beta$ (shifting factor):}
    \begin{itemize}
        \item \textbf{Higher $\beta$:} Emphasizes node-specific features, strengthening local feature representation in response to accidents.
        \item \textbf{Lower $\beta$:} Preserves global coherence, facilitating smoother information exchange across nodes.
    \end{itemize}
\end{itemize}
\(\gamma\) and \(\beta\) serve distinct yet complementary roles in guiding and adapting spatiotemporal modeling with Transformer. Their interplay enables the Transformer to balance global patterns with localized variations for robust performance across diverse scenarios.

\noindent\textbf{Condition Self-Attention.} Given the input embedding $\bm{X}^o$ and the contextual condition representation $\bm{X}^c$, GLN is applied to get the normalized input representation $\bm{X}^{(GLN)}$ through Eq.~(\ref{eq:GLN}) that adapts to the current traffic context. 
Building upon this, we extend the vanilla self-attention in Eq.~(\ref{eq:self-attention}) to a \textit{conditional} variant that incorporates both normalized input features and contextual condition information. Specifically, we override Eq.~(\ref{eq:self-attention}) as follows: 
\begin{equation}
\begin{aligned}
\bm{Q} &= \bm{X}^{(GLN)} \bm{W}_Q, \\ 
\bm{K} &= (\bm{X}^{(GLN)} \parallel \bm{X}^c) \bm{W}_K,  \\
\bm{V} &= (\bm{X}^{(GLN)} \parallel \bm{X}^c) \bm{W}_V.
\end{aligned}
\end{equation}
The subsequent spatial/temporal attention computations are identical to those in Eq.~(\ref{eq:st-attention}), allowing seamless integration of our conditional self-attention into the spatiotemporal Transformer backbone.

\noindent\textbf{Conditional Residual Connections.} 
Given the contextual condition representation $\bm{X}^c$, we propose learning a modulation factor $\alpha$ via an MLP, $\alpha = \texttt{MLP}(\bm{X}^c)$, to scale each module’s output through the residual connections. The design of this mechanism is informed by insights from \cite{goyal2017accurate}, where zero-initialization is leveraged to enhance the stability and efficiency of large-scale training processes. This approach has demonstrated significant benefits when applied to complex datasets like San Diego and Bay Area~\cite{LargeST}. From a conceptual perspective, this mechanism resonates with the principles of curriculum learning \cite{bengio2009curriculum}. By prioritizing the learning of fundamental, intrinsic features in the initial stages, the model progressively integrates additional contextual signals, which are typically noisier. This staged learning process improves optimization during training by enabling the model to focus on mastering essential patterns before adapting to more complex or variable information.


Building on the above components, we now present the whole workflow of our proposed ConFormer as follows: 
\begin{itemize}[leftmargin=0.4cm]
\item[$\bullet$] \textbf{Conditional Branch.} The conditional branch refines input features \(\bm{X}^o\) using graph structure \(\mathcal{A}\) through \texttt{GraphPropagation}, producing conditional features \(\bm{X}^c\). The MLP then outputs the conditional factors:
\begin{align}
	& \bm{X}^c = \texttt{GraphPropagation}(\bm{X}^o, \mathcal{A}), \\ 
	& [\gamma_c, \beta_c, \alpha_c] = \texttt{MLP}_c(\bm{X}^c).
\end{align}
Here, \(\beta_c\) and \(\gamma_c\) parameterize the mean and variance in GLN, while \(\alpha_c\) scales residual connections to enhance feature interactions. These factors adaptively influence the  transformer branch through context-aware attention and normalization.
\item[$\bullet$] \textbf{Transformer Branch.} The transformer branch incorporates the conditional factors as follows:
\begin{align}
	\bm{X}^{(GLN)} &= \texttt{GLN}(\bm{X}^o, \gamma_c, \beta_c), \\
	\bm{X}^{(Att)} &= \texttt{ConditionAttention}(\bm{X}^{(GLN)}, \bm{X}^c), \\
	\bm{X}^{(Res)} &= \bm{X}^o + \alpha_c \bm{X}^{(Att)}.
\end{align}
The \texttt{ConditionAttention} module enables dynamic adaptation of feature interactions, while the scaled residual connection \(\alpha_c\) ensures information continuity and stability.
\item[$\bullet$] \textbf{Feedforward Layer.} The feedforward module further processes the features using a secondary set of conditional factors \([\gamma_f, \beta_f, \alpha_f]\), also derived from \(\bm{X}^c\) via an MLP as follows:
\[
[\gamma_f, \beta_f, \alpha_f] = \texttt{MLP}_f(\bm{X}^c).
\]
These factors are applied to the normalized features through another GLN layer as follows:
\begin{align}
	\bm{X'}^{(GLN)} &= \texttt{GLN}(\bm{X}^{(Res)}, \gamma_f, \beta_f), \\
	\bm{X}^{(FF)} &= \texttt{FeedForward}(\bm{X'}^{(GLN)}), \\
	\bm{\hat{Y}}_t &=  \bm{X}^{(Res)} + \alpha_f \bm{X}^{(FF)}.
\end{align}
The residual connections, scaled by \(\alpha_f\), maintain model stability while facilitating deeper feature transformation. 
\end{itemize}

\begin{table}[b]
	\caption{Summary of Our Developed Datasets.}\label{tab:inf}
	\setlength{\abovecaptionskip}{0.cm}
	\setlength{\belowcaptionskip}{-0.0cm}
	\centering
		\resizebox{1\linewidth}{!}{\begin{tabular}{lccc}
		\hline Dataset & Tokyo & San Diego & Bay Area \\
		\hline Start Time & $2021 / 10 / 1$ & $2019 / 1 / 1$ & $2019 / 1 / 1$ \\
		End Time & $2021 / 12 / 31$ & $2019 / 12 / 31$ & $2019 / 12 / 31$ \\
		Time Interval & 10 minutes & 15 minutes & 15 minutes \\
		\#Spatial Units & 1,843 road links & 716 sensors  & 2,352 sensors \\
		Traffic Source & MegaCRN~\cite{MegaCRN} & LargeST~\cite{LargeST} & LargeST~\cite{LargeST} \\
		Accident Source & JARTIC\footnote{\url{https://www.jartic.or.jp/}} & US Accidents~\cite{moosavi2019countrywide} & US Accidents~\cite{moosavi2019countrywide} \\
		Regulation  & \ding{52} & \ding{55} & \ding{55} \\
		Accident & \ding{52} & \ding{52} & \ding{52} \\ 
		\hline
	\end{tabular}}
	\vspace{-5pt}
\end{table}

\begin{table*}[t]
\centering
\small
\renewcommand{\arraystretch}{1.1}
\setlength{\abovecaptionskip}{0.cm}
\setlength{\belowcaptionskip}{-0.0cm}
\tabcolsep=1mm
\caption{Performance comparison with improvements over the best baseline. All models incorporate incident data, for example, GWNet and AGCRN expand features, while STAEFormer and STID use binary embeddings. \textcolor{purple}{\textbf{Bold purple}} indicates the best results (ConFormer), \color{brown}\underline{brown underlined} \color{black} indicates the second-best results (best baseline).}

\label{tab:performance}
\begin{sc}
	\resizebox{\textwidth}{!}{
		\begin{tabular}{cccccc|ccc|ccc|ccc}
			\toprule
            			\multirow{2}{*}{Data} & \multirow{2}{*}{Method} & \multirow{2}{*}{Param} & \multicolumn{3}{c}{10 min} & \multicolumn{3}{c}{30 min} & \multicolumn{3}{c}{60 min} & \multicolumn{3}{c}{Average} \\ \cline{4-15} 
			&  &  & MAE & RMSE & MAPE & MAE & RMSE & MAPE & MAE & RMSE & MAPE & MAE & RMSE & MAPE \\ 
			\hline
			\multirowcell{11}{Tokyo\\($N = 1,843 $) \\(Accident \ding{52}) \\(Regulation \ding{52})} 
			& LSTM \cite{LSTM} & 98K & 7.00 & 11.10 & 31.95\% & 7.74 & 12.40 & 36.45\% & 8.65 & 13.81 & 41.81\% & 7.68 & 12.24 & 35.88\% \\
			& AGCRN \cite{AGCRN} & 771K & 5.99 & 9.48 & 25.71\% & 6.64 & 10.73 & 29.81\% & 6.99 & 11.39 & 32.13\% & 6.64 & 10.53 & 29.88\% \\
			& STGCN \cite{stgcn}  & 1.1M & 6.09 & 9.70 & 24.84\% & 6.91 & 11.09 & 30.24\% & 8.41 & 12.80 & 32.90\% & 7.24 & 11.20 & 29.99\% \\
			& GWNET \cite{GWNet} & 333K & 5.91 & 9.40 & 25.22\% & 6.59 & 10.64 & 29.78\% & 6.89 & 11.17 & 31.71\% & 6.56 & 10.40 & 29.90\% \\
			& STGODE \cite{STGODE} & 769K & 6.08 & 9.57 & 26.51\% & 6.74 & 10.75 & 30.29\% & 7.15 & 11.48 & 32.91\% & 6.79 & 10.64 & 30.87\% \\
			& STID \cite{STID} & 576K & 6.08 & 9.56 & 25.87\% & 6.85 & 10.90 & 31.25\% & 7.46 & 11.32 & 32.31\% & 6.60 & 10.51 & 29.81\% \\
			& DSTAGNN \cite{DSTAGNN} & 17.8M & 5.90 & 9.39 & 24.53\% & 6.68 & 10.87 & 29.93\% & 7.11 & 11.66 & 32.56\% & 6.67 & 10.64 & 29.34\% \\
			& DGCRN \cite{li2023dynamic} & 333K & 5.86 & 9.36 & 24.80\% & 6.49 & \color{brown}\underline{10.54} & \color{brown}\underline{29.23\%} & \color{brown}\underline{6.81} & \color{brown}\underline{11.11} & \color{brown}\underline{31.39\%} & \color{brown}\underline{6.49} & \color{brown}\underline{10.34} & 29.47\% \\
			& D$^2$STGNN \cite{D2STGNN} & 426K & \color{brown}\underline{5.85} & \color{brown}\underline{9.35} & \color{brown}\underline{24.52\%} & \color{brown}\underline{6.49} & 10.45 & 28.98\% & 6.83 & 11.12 & 31.12\% & 6.54 & 10.37 & \color{brown}\underline{29.44\%} \\
			& STAEFormer \cite{STAEformer} & 1.7M & 5.89 & 9.37 & 25.10\% & 6.52 & 10.52 & 29.00\% & 6.87 & 11.24 & 31.22\% & 6.56 & 10.39 & 29.84\% \\
			\hline
			& \multirow{2}{*}{\bf ConFormer} & \multirow{2}{*}{684K} &  \color{purple}\textbf{5.77} & \color{purple}\textbf{9.08} & \color{purple}\textbf{19.69\%} & \color{purple}\textbf{6.40} & \color{purple}\textbf{10.23} & \color{purple}\textbf{23.16\%} & \color{purple}\textbf{6.73} & \color{purple}\textbf{10.87} & \color{purple}\textbf{25.28\%} & \color{purple}\textbf{6.37} & \color{purple}\textbf{10.22} & \color{purple}\textbf{23.13\%} \\
			& & & \cellcolor{green!30}$\textbf{1.4\%}\uparrow$ & \cellcolor{green!30}$\textbf{2.9\%}\uparrow$ & \cellcolor{green!30}$\textbf{19.7\%}\uparrow$ & \cellcolor{green!30}$\textbf{1.4\%}\uparrow$ & \cellcolor{green!30}$\textbf{2.9\%}\uparrow$ & \cellcolor{green!30}$\textbf{20.8\%}\uparrow$ & \cellcolor{green!30}$\textbf{1.2\%}\uparrow$ & \cellcolor{green!30}$\textbf{2.2\%}\uparrow$ & \cellcolor{green!30}$\textbf{19.5\%}\uparrow$ & \cellcolor{green!30}$\textbf{1.7\%}\uparrow$ & \cellcolor{green!30}$\textbf{1.2\%}\uparrow$ & \cellcolor{green!30}$\textbf{21.5\%}\uparrow$ \\
            \bottomrule
			\multirow{2}{*}{Data} & \multirow{2}{*}{Method} & \multirow{2}{*}{Param} & \multicolumn{3}{c}{45 min} & \multicolumn{3}{c}{90 min} & \multicolumn{3}{c}{180 min} & \multicolumn{3}{c}{Average} \\ \cline{4-15}
			&  &  & MAE & RMSE & MAPE & MAE & RMSE & MAPE & MAE & RMSE & MAPE & MAE & RMSE & MAPE \\
			\midrule
			\multirowcell{13}{San Diego\\(\text{$N = 716$}) \\(Accident \ding{52}) }
			& LSTM  \cite{LSTM} & 98K & 19.03 & 30.53 & 11.81\% & 25.84 & 40.87 & 16.44\% & 37.63 & 59.07 & 25.45\% & 26.44 & 41.73 & 17.20\% \\
			& AGCRN \cite{AGCRN} & 761K & 15.71 & 27.85 & 11.48\% & 18.06 & 31.51 & 13.06\% & {21.86} & 39.44 & 16.52\% & 18.09 & 32.01 & 13.28\% \\
			& STGCN \cite{stgcn}  & 508K & 17.45 & 29.99 & 12.42\% & 19.55 & 33.69 & 13.68\% & 23.21 & 41.23 & 16.32\% & 19.67 & 34.14 & 13.86\% \\
			& GWNET \cite{GWNet} & 311K & 15.35 & {25.17} & 10.67\% & 18.23 & 30.13 & 12.21\% & {22.44} & {38.02} & 15.69\% & {18.14} & 30.11 & 12.27\% \\
			& STGODE \cite{STGODE} & 729K & 16.75 & 28.04 & 11.00\% & 19.71 & 33.56 & 13.16\% & 23.67 & 42.12 & 16.58\% & 19.55 & 33.57 & 13.22\% \\
			& DSTAGNN \cite{DSTAGNN} & 3.9M & 18.13 & 28.96 & 11.38\% & 21.71 & 34.44 & 13.93\% & 27.51 & 43.95 & 19.34\% & 21.82 & 34.68 & 14.40\% \\
			& STID \cite{STID} & 258K & \color{brown}\underline{15.14} & \color{brown}\underline{25.07} & \color{brown}\underline{9.84\%} & \color{brown}\underline{17.63} & \color{brown}\underline{29.16} & \color{brown}\underline{11.46\%} & \color{brown}\underline{21.02} & \color{brown}\underline{36.72} & \color{brown}\underline{15.02\%} & \color{brown}\underline{17.57} & \color{brown}\underline{28.92} & \color{brown}\underline{11.45\%} \\
			& DGCRN \cite{li2023dynamic} & 243K & {15.34} & 25.35 & {10.01\%} & 18.05 & 30.06 & {11.90\%} & 22.06 & 37.51 & 15.27\% & 18.02 & 30.09 & {12.07\%} \\
			& STAEFormer \cite{STAEformer} & 1.7M & 15.52 & 25.78 & 10.21\% & 18.23 & 30.66 & 12.20\% & 22.39 & 37.88 & 15.60\% & 18.21 & 30.44 & 12.27\% \\
			& D$^2$STGNN \cite{D2STGNN} & 406K & 15.58 & 25.74 & 10.55\% & {18.00} & {29.98} & 12.34\% & 21.96 & {37.10} & {15.06\%} & {17.96} & {29.97} & 12.11\% \\
			\hline
			&\multirow{2}{*}{\textbf{ConFormer}} & \multirow{2}{*}{758K} &  \color{purple}\textbf{14.52} & \color{purple}\textbf{24.33} & \color{purple}\textbf{9.24\%} & \color{purple}\textbf{16.82} & \color{purple}\textbf{28.37} & \color{purple}\textbf{10.81\%} & \color{purple}\textbf{20.18} & \color{purple}\textbf{34.53} & \color{purple}\textbf{13.52\%} & \color{purple}\textbf{16.75} & \color{purple}\textbf{28.59} & \color{purple}\textbf{10.88\%} \\
			& & & \cellcolor{green!30}{$\textbf{4.1\%}\uparrow$} & \cellcolor{green!30}{$\textbf{2.9\%}\uparrow$} & \cellcolor{green!30}{$\textbf{6.1\%}\uparrow$} & \cellcolor{green!30}{$\textbf{4.6\%}\uparrow$} & \cellcolor{green!30}{$\textbf{2.7\%}\uparrow$} & \cellcolor{green!30}{$\textbf{5.7\%}\uparrow$} & \cellcolor{green!30}{$\textbf{4.0\%}\uparrow$} & \cellcolor{green!30}{$\textbf{6.0\%}\uparrow$} & \cellcolor{green!30}{$\textbf{10.0\%}\uparrow$} & \cellcolor{green!30}{$\textbf{4.7\%}\uparrow$} & \cellcolor{green!30}{$\textbf{1.1\%}\uparrow$} & \cellcolor{green!30}{$\textbf{5.0\%}\uparrow$} \\
			\bottomrule
			\multirowcell{11}{Bay Area\\($N = 2,352$)\\(Accident \ding{52})}
			& LSTM \cite{LSTM} & 98K & 20.38 & 33.34 & 15.47\% & 27.56 & 43.57 & 23.52\% & 39.03 & 60.59 & 37.48\% & 27.96 & 44.21 & 24.48\% \\
			& AGCRN \cite{AGCRN} & 777K & 18.31 & 30.24 & 14.27\% & 21.27 & 34.72 & 16.89\% & 25.85 & 40.18 & 20.80\% & 21.01 & 34.25 & 16.90\% \\
			& STGCN \cite{stgcn}  & 1.3M & 21.05 & 34.51 & 16.42\% & 23.63 & 38.92 & 18.35\% & 26.87 & 44.45 & 21.92\% & 23.42 & 38.57 & 18.46\% \\
			& GWNET \cite{GWNet} & 344K & 17.85 & 29.12 & 13.92\% & 21.11 & 33.69 & 17.79\% & 25.58 & 40.19 & 23.48\% & 20.91 & 33.41 & 17.66\% \\
			& STGODE \cite{STGODE} & 788K & 18.84 & 30.51 & 15.43\% & 22.04 & 35.61 & 18.42\% & 26.22 & 42.90 & 22.83\% & 21.79 & 35.37 & 18.26\% \\
			& DSTAGNN \cite{DSTAGNN}  & 26.9M & 19.73 & 31.39 & 15.42\% & 24.21 & 37.70 & 20.99\% & 30.12 & 46.40 & 28.16\% & 23.82 & 37.29 & 20.16\% \\
			& DGCRN \cite{li2023dynamic} & 374K & 18.02 & 29.49 & 14.13\% &  21.08 & 34.03 & 16.94\% & 25.25 & 40.63 & 21.15\% & 20.91 & 33.83 & 16.88\% \\
			& STID \cite{STID} & 711K & \color{brown}\underline{17.30} & \color{brown}\underline{29.02} & 13.56\% & \color{brown}\underline{20.47} & \color{brown}\underline{33.34}  & 16.60\% & \color{brown}\underline{24.35} & \color{brown}\underline{39.87} & 20.60\% & \color{brown}\underline{20.19} & 33.47 & 16.36\% \\
			& D$^2$STGNN \cite{D2STGNN} & 446K & 17.54 & 28.94 & \color{brown}\underline{12.12\%} & 20.92 & 33.92 & \color{brown}\underline{15.29\%} & 25.48 & 40.99 & \color{brown}\underline{19.83\%} & 20.71 & 33.65 & \color{brown}\underline{15.34\%} \\
			& STAEFormer \cite{STAEformer} & 1.8M & 17.58 & 29.04 & 12.59\% & 21.04 & 34.22 & 16.65\% & 25.86 & 40.52 & 20.42\% & 20.80 & \color{brown}\underline{33.37} & 15.44\% \\
			\hline
			& \multirow{2}{*}{\bf ConFormer} & \multirow{2}{*}{784K} &  \color{purple}\textbf{16.93} & \color{purple}\textbf{28.48} & \color{purple}\textbf{12.09\%} & \color{purple}\textbf{19.96} & \color{purple}\textbf{32.89} & \color{purple}\textbf{15.15\%} & \color{purple}\textbf{24.03} & \color{purple}\textbf{39.06} & \color{purple}\textbf{19.68\%} & \color{purple}\textbf{19.82} & \color{purple}\textbf{33.00} & \color{purple}\textbf{15.24\%} \\
			& & & \cellcolor{green!30}$\textbf{2.1\%}\uparrow$ & \cellcolor{green!30}$\textbf{1.9\%}\uparrow$ & \cellcolor{green!30}$\textbf{0.2\%}\uparrow$ & \cellcolor{green!30}$\textbf{5.3\%}\uparrow$ & \cellcolor{green!30}$\textbf{3.4\%}\uparrow$ & \cellcolor{green!30}$\textbf{10.6\%}\uparrow$ & \cellcolor{green!30}$\textbf{4.8\%}\uparrow$ & \cellcolor{green!30}$\textbf{3.9\%}\uparrow$ & \cellcolor{green!30}$\textbf{7.0\%}\uparrow$ & \cellcolor{green!30}$\textbf{1.8\%}\uparrow$ & \cellcolor{green!30}$\textbf{1.4\%}\uparrow$ & \cellcolor{green!30}$\textbf{0.7\%}\uparrow$ \\
			\bottomrule
	\end{tabular}}
\end{sc}
\end{table*}

\subsection{Computational Complexity}\label{sec:cost}
We partition the computational complexity of ConFormer into three primary components and calculate the FLOPs for each as follows:

\begin{itemize}[leftmargin=0.5cm]
\item[$\bullet$] \textbf{Graph Propagation.} Graph propagation leverages the Chebyshev polynomial approximation with a computational complexity of $\mathcal{O}(K|\mathcal{E}|D)$, where $K$ denotes the order of the polynomial, $|\mathcal{E}|$ represents the number of edges, and $D$ is the number of channels. This complexity primarily arises from the approximation of the graph Laplacian operator.

\item[$\bullet$] \textbf{Condition Self-Attention.} The complexity of spatiotemporal condition self-attention is $\mathcal{O}(TN^2D + NT^2D)$, where $N$ is the number of spatial nodes and $T$ is the temporal length.

\item[$\bullet$] \textbf{GLN.} The generation of $\gamma$, $\beta$, and $\alpha$ for layer normalization and residual connections incurs a computational cost of $NTD^2$.
\end{itemize}
Given that ConFormer is largely controlled by conditions, a single attention layer suffices to achieve superior performance. Therefore, the total FLOPs of ConFormer can be computed as:
\begin{align*}
\text{FLOPs(ConFormer)} = K|\mathcal{E}|D + (TN^2D + NT^2D) + NTD^2.
\end{align*}


\begin{table*}[t]
\centering
\caption{Performance on the standard benchmarks, i.e., without using any accident information.}
\label{tab:perf}
\setlength{\abovecaptionskip}{0.cm}
\setlength{\belowcaptionskip}{-0.0cm}
\renewcommand\arraystretch{1.2}
\begin{sc}
	\resizebox{\linewidth}{!}{%
		\begin{tabular}{ccc|cccccccccc|c}
			\toprule
			\multicolumn{2}{c}{Dataset} & Metric & HI & STGCN \cite{stgcn}  & DCRNN \cite{DCRNN} & GWNet \cite{GWNet} & AGCRN \cite{AGCRN} & GTS \cite{GTS} & STID \cite{STID} & PDFormer \cite{PDFormer} & STAEFormer \cite{STAEformer}& ConFormer & Improve \\
			\hline
			\multirow{3}{*}{\rotatebox{90}{PEMS03}} & \multirow{3}{*}{Average} & MAE & 32.62 & 15.83 & 15.54 & \color{brown}\underline{14.59} & 15.24 & 15.41 & 15.33 & 14.94 & 15.35 & \color{purple}\textbf{14.51} & \cellcolor{green!30}$\textbf{5.5\%}\uparrow$ \\
			& & RMSE & 49.89 & 27.51 & 27.18 & \color{brown}\underline{25.24} & 26.65 & 26.15 & 27.40 & 25.39 & 27.55 & \color{purple}\textbf{24.81} & \cellcolor{green!30}$\textbf{2.3\%}\uparrow$ \\
			& & MAPE & 30.60\% & 16.13\% & 15.62\% & 15.52\% & 15.89\% & 15.39\% & 16.40\% & 15.82\% & \color{brown}\underline{15.18\%} & \color{purple}\textbf{14.38\%} & \cellcolor{green!30}$\textbf{5.3\%}\uparrow$ \\
			\midrule
			\multirow{3}{*}{\rotatebox{90}{PEMS04}} & \multirow{3}{*}{Average} & MAE & 42.35 & 19.57 & 19.63 & 18.53 & 19.38 & 20.96 & 18.38 & 18.36 & \color{brown}\underline{18.22} & \color{purple}\textbf{17.89} & \cellcolor{green!30}$\textbf{1.8\%}\uparrow$ \\
			& & RMSE & 61.66 & 31.38 & 31.26 & \color{brown}\underline{29.92} & 31.25 & 32.95 & 29.95 & 30.03 & 30.18 & \color{purple}\textbf{29.79} & \cellcolor{green!30}$\textbf{1.3\%}\uparrow$ \\
			& & MAPE & 29.92\% & 13.44\% & 13.59\% & 12.89\% & 13.40\% & 14.66\% & 12.04\% & 12.00\% & \color{brown}\underline{11.98\%} & \color{purple}\textbf{11.83\%} & \cellcolor{green!30}$\textbf{1.3\%}\uparrow$ \\
			\midrule
			\multirow{3}{*}{\rotatebox{90}{PEMS07}} & \multirow{3}{*}{Average} & MAE & 49.03 & 21.74 & 21.16 & 20.47 & 20.57 & 22.15 & 19.61 & 19.97 & \color{brown}\underline{19.14} & \color{purple}\textbf{19.04} & \cellcolor{green!30}$\textbf{0.5\%}\uparrow$ \\
			& & RMSE & 71.18 & 35.27 & 34.14 & 33.47 & 34.40 & 35.10 & 32.79 & 32.95 & \color{brown}\underline{32.60} & \color{purple}\textbf{32.50} & \cellcolor{green!30}$\textbf{0.3\%}\uparrow$ \\
			& & MAPE & 22.75\% & 9.24\% & 9.02\% & 8.61\% & 8.74\% & 9.38\% & 8.30\% & 8.55\% & \color{brown}\underline{8.01\%} & \color{purple}\textbf{7.89\%} & \cellcolor{green!30}$\textbf{1.5\%}\uparrow$ \\
			\midrule
			\multirow{3}{*}{\rotatebox{90}{PEMS08}} & \multirow{3}{*}{Average} & MAE & 36.66 & 16.08 & 15.22 & 14.40 & 15.32 & 16.49 & 14.21 & 13.58 & \color{brown}\underline{13.46} & \color{purple}\textbf{13.32} & \cellcolor{green!30}$\textbf{1.0\%}\uparrow$ \\
			& & RMSE & 50.45 & 25.39 & 24.17 & 23.39 & 24.41 & 26.08 & 23.28 & 23.41 & \color{brown}\underline{23.25} & \color{purple}\textbf{22.96} & \cellcolor{green!30}$\textbf{1.2\%}\uparrow$ \\
			& & MAPE & 21.63\% & 10.60\% & 10.21\% & 9.21\% & 10.03\% & 10.54\% & 9.27\% & 9.05\% & \color{brown}\underline{8.88\%} & \color{purple}\textbf{8.76\%} & \cellcolor{green!30}$\textbf{1.4\%}\uparrow$ \\
			\bottomrule
		\end{tabular}%
	}
\end{sc}
\end{table*}

\section{Experiment}
\subsection{Experimental Setup}\label{sec:data}
\noindent\textbf{Datasets.} We collect large-scale traffic datasets along with accident and regulation information from both Tokyo and California. Details of the datasets are summarized in Table \ref{tab:inf}. The traffic accidents for both regions are visualized in Figure~\ref{fig:1b}. \textit{We will release these datasets once the article is accepted.}

\begin{itemize}[leftmargin=0.4cm]
\item First, we develop a comprehensive dataset encompassing 1,843 Tokyo expressway links over three months (October–December 2021). This dataset advances beyond existing collections \cite{MegaCRN} by incorporating authoritative incident data from the Japan Road Traffic Information Center (JARTIC) \cite{jartic2025}, maintained by Japan's Ministry of Land, Infrastructure, Transport and Tourism. Our integration yields unprecedented 10-minute synchronized observations of traffic speeds, accidents, and regulatory interventions. The dataset captures fine-grained regulatory categories (single-lane restrictions, double-lane restrictions, road closures) and comprehensive accident classifications by severity (property damage, injury, fatality) and causative factors (overturns, collisions, fires, cargo collapse, debris, construction disruptions).

\item Next, we systematically enhance California highway traffic data in San Diego (716 stations) and Bay Area (2,352 stations) published by LargeST ~\cite{LargeST}, by incorporating the comprehensive U.S. Accidents dataset \cite{moosavi2019countrywide}, containing 1.5 million georeferenced incident records across 49 states. These records, continuously collected since February 2016 from authoritative sources including MapQuest Traffic \cite{mapquest2019} and Microsoft Bing Map Traffic \cite{bing2019}, provide detailed severity assessments from transportation departments, law enforcement, and traffic monitoring infrastructure.
\end{itemize}

\noindent\textbf{Settings.} Experiments were conducted on a GPU server with eight GeForce GTX 3090 GPUs using PyTorch 2.0.3. Data was standardized via z-score normalization~\cite{cheadle2003analysis}. Training employed early stopping, terminating if validation error stabilized within 15–20 epochs or showed no improvement after 200 epochs, with the best model retained based on validation performance~\cite{luo2023dynamic}. Data was split chronologically into training, validation, and test sets in a 6:2:2 ratio. Model performance was evaluated using Mask-Based RMSE, MAE, and MAPE, excluding zero values as noise.

\begin{table*}[!htbp]
	\renewcommand{\arraystretch}{1.1}
	\setlength{\abovecaptionskip}{0.cm}
	\setlength{\belowcaptionskip}{-0.0cm}
	\caption{Comprehensive Ablation Studies}
	\label{tab:ablation_completed}
	\centering
	\setlength{\tabcolsep}{4pt}
	\resizebox{1.8\columnwidth}{!}{
		\begin{tabular}{c|c|c|c|ccccccc}
			\toprule
			\multirow{2}{*}{Variation} &\multicolumn{3}{c|}{Component} & \multicolumn{3}{c}{Tokyo} & \multicolumn{2}{c}{San Diego} & \multicolumn{2}{c}{Bay Area} \\
			\cmidrule(lr){2-11}  
			& Spatial & Temporal & LN & Normal & Accident & Regulation & Normal & Accident & Normal & Accident \\
			\midrule
			\textbf{ConFormer} & \cellcolor{green!30}\textbf{Attention} & \cellcolor{green!30}\textbf{Attention} & \cellcolor{green!30}\textbf{GLN} & \cellcolor{green!30}\textbf{6.37} & \cellcolor{green!30}\textbf{8.98} & \cellcolor{green!30}\textbf{9.18} & \cellcolor{green!30}\textbf{16.75} & \cellcolor{green!30}\textbf{22.36} & \cellcolor{green!30}\textbf{19.82} & \cellcolor{green!30}\textbf{26.95} \\
			
			\midrule
			\multirow{7}{*}{Module} & Attention & Attention & w/ LN & 6.45 & 9.58 & 9.30 & 16.86 & 23.49 & 19.96 & 28.45 \\
			& w/ MLP & w/ MLP & GLN & 6.75 & 10.06 & 9.42 & 17.39 & 24.19 & 20.57 & 29.30 \\
			\cmidrule(lr){2-11}
			& Attention & w/o & GLN & 6.55 & 9.34 & 9.53 & 17.10 & 22.81 & 20.23 & 27.49 \\
			& w/o & Attention & GLN & 6.75 & 9.42 & 9.65 & 17.16 & 23.01 & 20.27 & 27.55 \\
			\cmidrule(lr){2-11}  
			& Attention & Attention & w/o $\alpha$ & 6.48 & 9.15 & 9.53 & 16.95 & 22.45 & 19.99 & 27.08 \\
			& Attention & Attention & w/o $\beta$ &  6.44 & 9.33 & 9.53 & 16.81 & 23.68 & 19.92 & 28.15 \\
			& Attention & Attention & w/o $\gamma$ & 6.46 & 9.42 & 9.53 & 16.80 & 23.49 & 19.96 & 28.33 \\
			\midrule
			\multirow{2}{*}{Data} & \multicolumn{2}{c|}{w/o Accident} & GLN  & 6.39 & 10.97 & 10.12 & 16.81 & 23.27 & 19.87 & 28.17 \\
			& \multicolumn{2}{c|}{w/o Regulation} & GLN & 6.41 & 9.48 & 10.64 & - & - & - & - \\
			
			\bottomrule
	\end{tabular}}
\end{table*}

\begin{table}[h]
	\centering
	\setlength{\abovecaptionskip}{0.1cm}
	\setlength{\belowcaptionskip}{-0.0cm}
	\caption{Performance of GLN on GWNet}
	\resizebox{0.95\columnwidth}{!}{\begin{tabular}{l|cc|cc|cc}
			\hline
			\multirow{2}{*}{Model} & \multicolumn{2}{c|}{Tokyo} & \multicolumn{2}{c|}{San Diego} & \multicolumn{2}{c}{Bay Area} \\
			& MAE & RMSE & MAE & RMSE & MAE & RMSE \\
			\hline
			GWNet & 20.91 & 33.41 & 18.14 & 30.11 & 6.56 & 10.40 \\
			GLN+GWNet & \cellcolor{green!30}\textbf{20.25} & \cellcolor{green!30}\textbf{32.74} &\cellcolor{green!30} \textbf{17.64} & \cellcolor{green!30}\textbf{29.45} & \cellcolor{green!30}\textbf{6.43} & \cellcolor{green!30}\textbf{10.21} \\
			\hline
	\end{tabular}}
	\vspace{-5pt}
	\label{tab:gwnet-comparison}
\end{table}
\subsection{Performance Evaluation}
\textbf{Baselines.} We included the following representative baselines in our study: Historical Inertia (HI)~\cite{liang2021revisiting}, a naive method that uses the average of historical values as the prediction result; LSTM~\cite{LSTM}, solely focusing on temporal aspects and ignoring spatial correlations; GNN-based models including DCRNN~\cite{DCRNN}, AGCRN~\cite{AGCRN}, STGCN~\cite{stgcn}, GWNET~\cite{GWNet}, DSTAGNN~\cite{DSTAGNN}, DGCRN~\cite{li2023dynamic}, and D$^2$STGNN~\cite{D2STGNN}; Attention-based models including ASTGCN~\cite{ASTGCN} and STTN~\cite{xu2020spatial}; Transformer variants such as PDFormer~\cite{PDFormer} and STAEFormer~\cite{STAEformer}; STGODE~\cite{STGODE}, a controlled differential equation-based method to model continuous traffic signal changes.

\noindent\textbf{Performance on Our Datasets.} \tableref{tab:performance} reports MAE, RMSE, and MAPE results for the Tokyo, Bay Area, and San Diego datasets across multiple prediction horizons, along with average performance. For the Tokyo dataset, following \cite{MegaCRN}, the input and output lengths are set to $T = 6$ and $T' = 6$. For the Bay Area and San Diego datasets, following \cite{LargeST}, the input and output lengths are $T = 12$ and $T' = 12$. 
GWNET combines GNN and Gated TCN, while DCRNN incorporates diffusion convolution into GRU. DGCRN extends DCRNN by modeling dynamic spatial topologies, achieving superior results among baselines. STAEFormer excels in long-term predictions due to its ability to capture extended dependencies via attention mechanisms but faces scalability challenges on larger datasets like Bay Area. 
ConFormer achieves SOTA performance across all datasets and prediction horizons, outperforming GCN-based methods (e.g., D$^2$STGNN, GWNET) and Transformer-based approaches (e.g., PDFormer, STAEFormer). The integration of global traffic context and conditional layer normalization contributes to its superior results. Crucially, ConFormer is the first Transformer-based model capable of operating on very large graphs, demonstrating significant improvements and establishing Transformer's scalability and effectiveness for large-scale traffic prediction.

\noindent\textbf{Performance on Benchmarks.} To further verify the performance of our method, we also conducted experiments on the widely-used benchmarks~\cite{ASTGCN, DLTraff} with input and output lengths are set to $T = 12$ and $T' = 12$.  It is important to note that these benchmark datasets (PEMS03/04/07/08) do not contain traffic accident or regulation information, thus our ConFormer model operates without these additional contextual inputs in these experiments.
Table~\ref{tab:perf} presents the experimental results. ConFormer consistently achieved the best performance across all ranges and datasets, demonstrating the effectiveness of our model. \textit{The superior performance of ConFormer on these traditional benchmarks demonstrates that our model can effectively utilize traffic flow signals as proxy variables for underlying traffic conditions through graph propagation, where the GLN mechanism and conditional attention architecture adaptively capture implicit disruptions and anomalies embedded within the spatiotemporal traffic patterns, even without explicit incident annotations.}  The GLN mechanism enables dynamic adaptation to subtle variations in traffic patterns that may indicate implicit disruptions, while the conditional attention architecture learns to distinguish between regular flow fluctuations and anomalous conditions embedded within the spatiotemporal patterns. Through graph propagation, local traffic anomalies propagate across the network as measurable changes in flow characteristics, allowing ConFormer to capture these implicit incident signatures even without explicit annotations.

\subsection{Ablation Study}
Through comprehensive ablation studies, we systematically evaluated the contribution of each component in ConFormer. The baseline ConFormer, integrating spatial attention, temporal attention, and GLN, consistently achieves optimal performance across all datasets and scenarios. Detailed ablation experiments were conducted from four aspects:
(1) Replacing GLN with standard layer normalization leads to notable performance degradation (MAE increases from 8.98 to 9.58 in accident scenarios), while substituting attention mechanisms with MLPs results in more significant deterioration (MAE rises to 10.06), validating the necessity of our attention-based architecture.
(2) Further experiments removing individual components (\textit{w/o}) show that both spatial and temporal attention are crucial. Removing either causes performance drops across all datasets. For example, on the Tokyo dataset, removing temporal attention increases MAE from 6.37 to 6.55, and removing spatial attention raises it to 6.75.
(3) Each modulation factor ($\alpha$, $\beta$, $\gamma$) in GLN plays a distinct role in enhancing model performance:
\begin{itemize}[leftmargin=0.4cm]
	\item $\alpha$ primarily affects model optimization, influencing performance in both normal and accident scenarios (removing it increases MAE from 16.75 to 16.95 on San Diego dataset).
	\item $\beta$ increases model sensitivity to sudden traffic disruptions (its removal leads to MAE increase from 22.36 to 23.68 in accident scenarios).
	\item $\gamma$ captures node-specific conditions to distinguish between normal and accident scenarios (removing it increases MAE from 26.95 to 28.33 in Bay Area accident cases).
\end{itemize}
(4) we evaluated the importance of incorporating accident and regulation data. Removing accident information substantially impairs prediction accuracy in corresponding scenarios (MAE increases from 8.98 to 10.97 in Tokyo accident scenarios). Similarly, removing regulation data leads to worse performance in regulation scenarios (MAE increases from 9.18 to 10.64). This empirically validates the effectiveness of incorporating traffic incident information.

Besides, to validate the extensibility of our method, we implemented a condition-informed variant of GWNet, denoted as GLN+GWNet, and evaluated it across three datasets. As presented in Table \ref{tab:gwnet-comparison}, notably, GLN+GWNet achieved a 2.63\% reduction in MAE compared to the original GWNet across all datasets.

\subsection{Efficiency Evaluation}
\begin{figure}[t]
	\centering
	\includegraphics[width=1\linewidth]{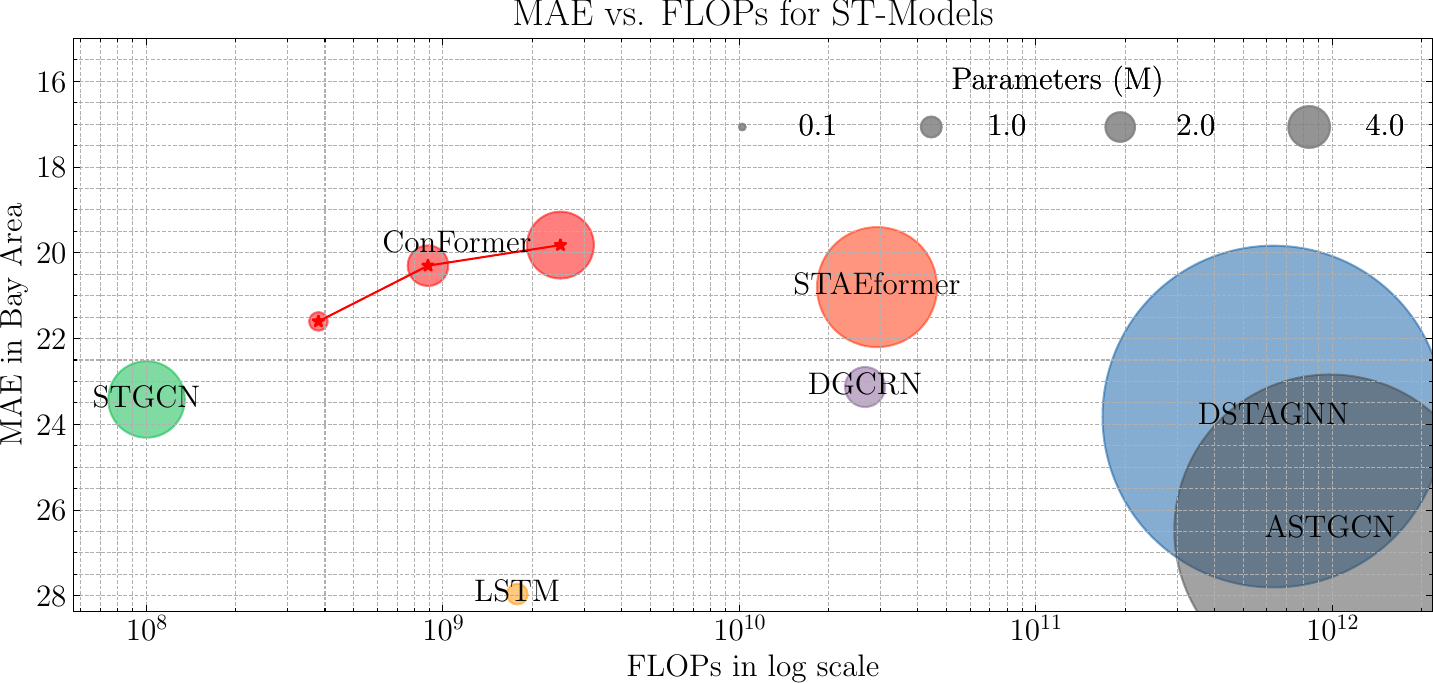}
	\caption{ Comparison of model efficiency and effectiveness across spatiotemporal models on the Bay Area dataset.  }
	\label{fig:bay_performance}
    \vspace{-15pt}
\end{figure}

\figref{fig:bay_performance} compares the efficiency and effectiveness of various spatiotemporal models on the Bay Area dataset. The x-axis shows computational complexity (FLOPs) on a logarithmic scale, the y-axis denotes predictive accuracy (MAE, with lower values indicating better performance), and circle size denotes model parameters (in millions, M). 
ConFormer demonstrates a notable balance between accuracy and computational cost, with MAE improving clearly as model complexity increases. STGCN achieves relatively low MAE with minimal computational overhead, while LSTM has the lowest complexity but poor accuracy. In contrast, ConFormer stands out by efficiently enhancing prediction performance with moderate increases in complexity. DSTAGNN and ASTGCN, however, achieve competitive accuracy at the cost of significantly higher computational demands and parameter counts. 


\begin{figure*}[t]
	\centering
	\includegraphics[width=0.48\linewidth]{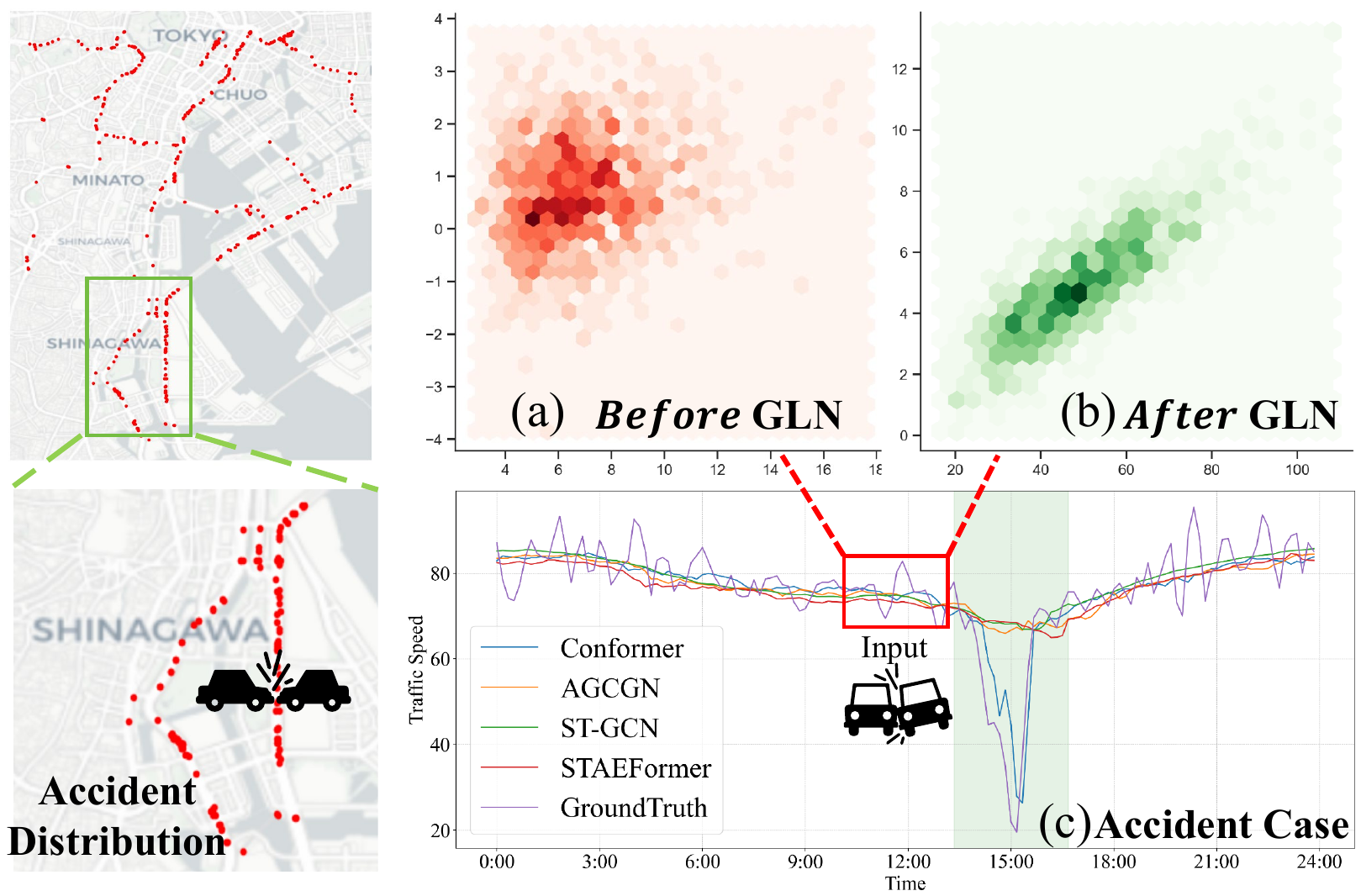}
	\includegraphics[width=0.48\linewidth]{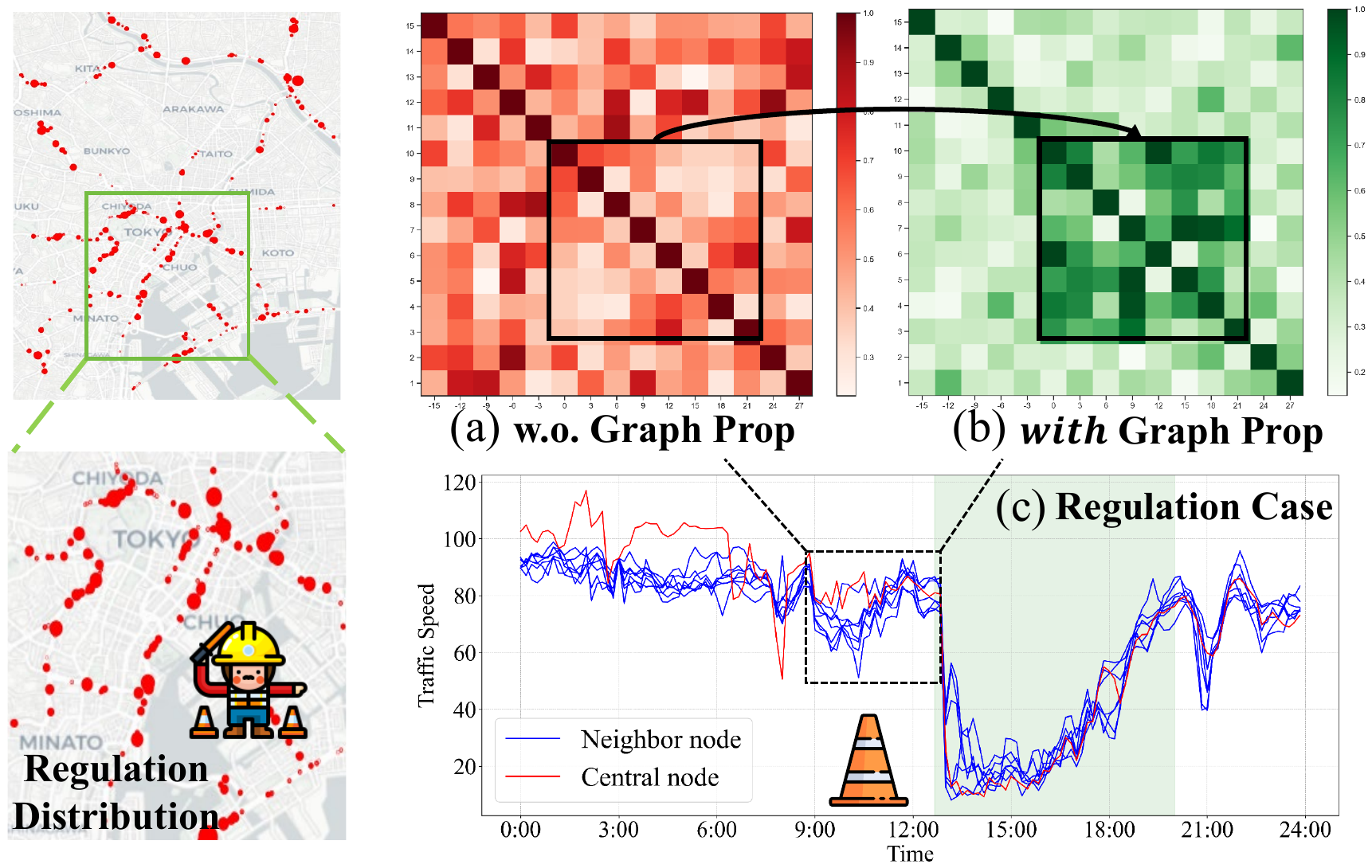}
	\caption{Case studies of traffic prediction in a subarea of Tokyo. Left: Analysis of an accident scenario, where ConFormer effectively captures the evolution of node distributions in latent space before and after the accident. Right: Analysis of a traffic regulation scenario, where graph propagation demonstrates learned differential attention scores in response to the implemented traffic regulations.}
	\label{fig:case12}
	\vspace{-10pt}
\end{figure*}

\begin{figure}[h]
	\subfloat[Normal]{\includegraphics[width=0.48\linewidth]{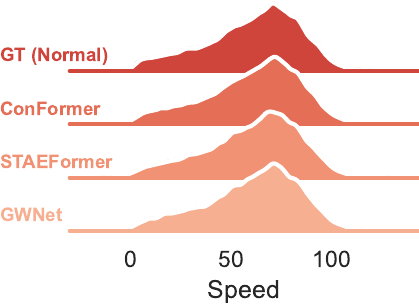}}
	\hfill
	\subfloat[Accident]{\includegraphics[width=0.48\linewidth]{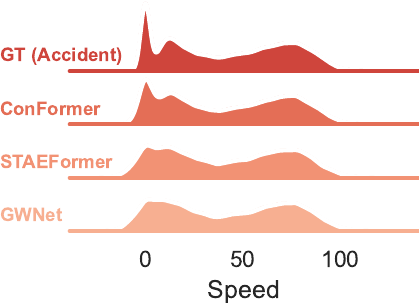}}
	\vspace{-0.5em}
	\caption{Distribution of predictions and ground-truths under normal and accident scenarios on Tokyo dataset.}
	\label{fig:compare}
\end{figure}

\begin{figure}[h]
	\centering
	\subfloat[Accident Case 1]{\includegraphics[width=0.48\linewidth]{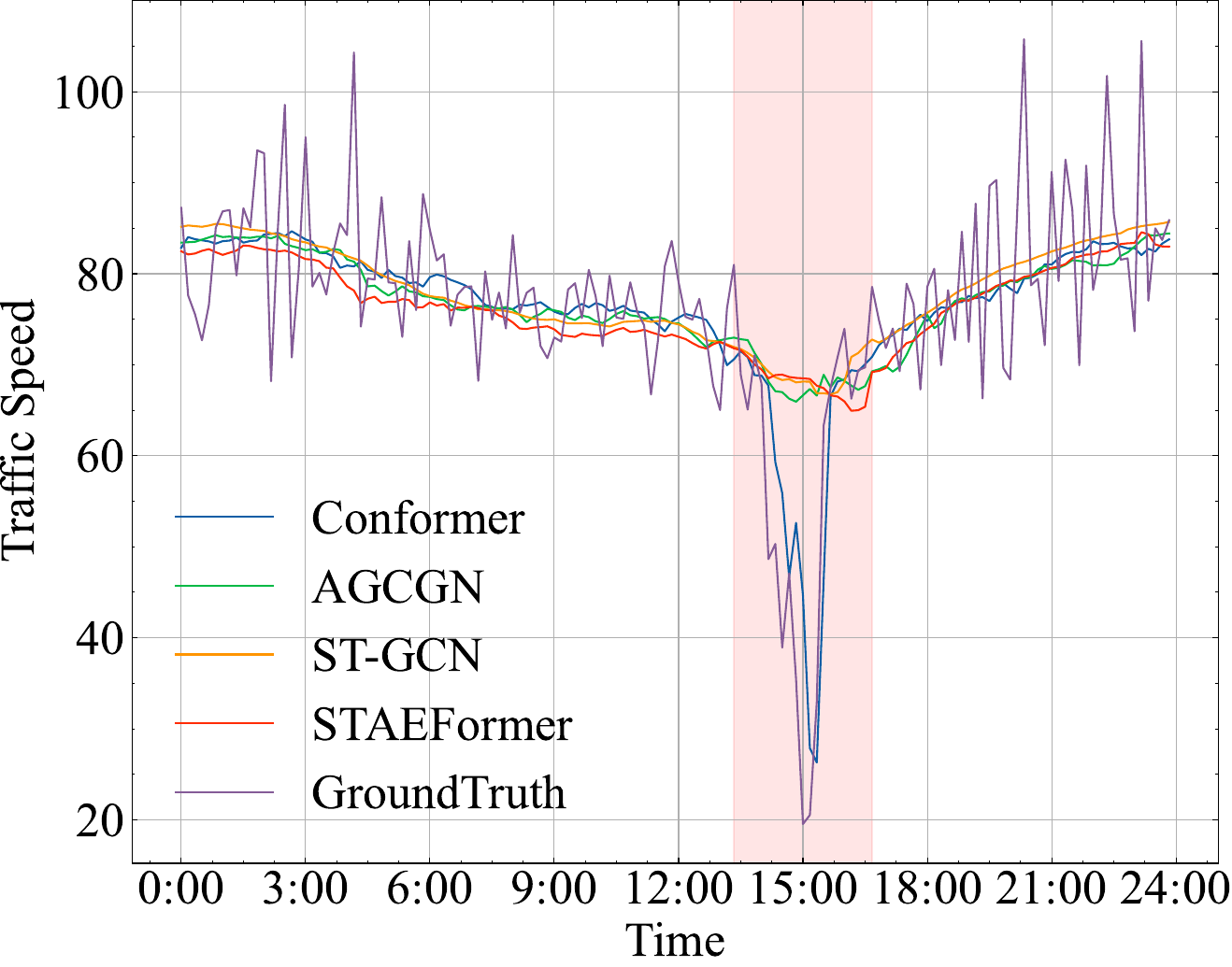}\label{app:case1}}
	\hfill
	\subfloat[Accident Case 2]{\includegraphics[width=0.48\linewidth]{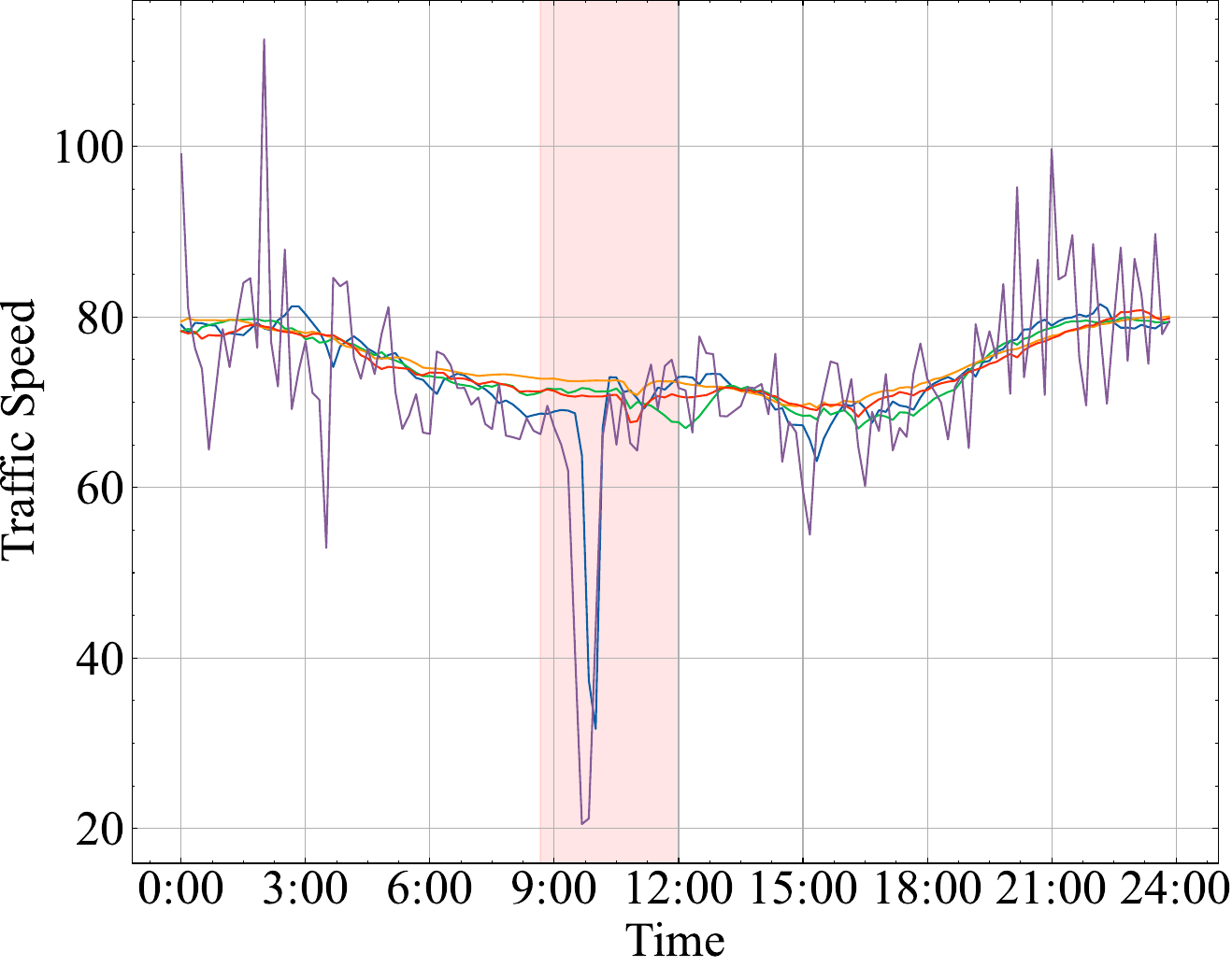}\label{app:case2}}
	\caption{Prediction results of two accident cases on Tokyo.}
	\label{app:case_study}
\end{figure}

\subsection{Case Study}
Using the Tokyo dataset, we conducted several specific case analyses: (1) For the \textbf{traffic accident scenario} (\figref{fig:case12}-left), our analysis showed that ConFormer effectively captures the evolving patterns of node distributions in the latent space. Using t-SNE dimensionality reduction, we visualized the feature spaces before and after the accident—before the accident, traffic time series showed normal, uncongested patterns, with tightly linked features among neighboring nodes. This hinders traditional models' ability to predict accidents; however, ConFormer uses the GLN mechanism with parameters $\gamma_{c}$ and $\beta$ to induce significant feature shifts, enabling more accurate differentiation of abnormal traffic conditions. (2) For the \textbf{traffic regulation scenario} (\figref{fig:case12}-right), we also tested the model's ability to transmit information and adapt to traffic changes. The central node implemented traffic regulations from 13:00 to 20:00: before the regulations, there was a clear difference in attention distribution between the central node and its neighbors; after the regulations began, the speeds of these nodes became highly consistent. This highlights the critical role of graph propagation in helping the model adapt to and transmit traffic information—through conditional mapping message passing, the attention maps of the central node and its neighbors show significantly enhanced correlation, demonstrating ConFormer's ability to dynamically respond to changes in the traffic environment.

As shown in \figref{fig:compare}, we compare ConFormer, STAEFormer, and GWNet predictions against ground-truth distributions for normal and accident scenarios on the Tokyo dataset. ConFormer demonstrates superior performance in accident scenarios while maintaining comparable accuracy in normal conditions, particularly excelling at handling sudden traffic disruptions.

\figref{app:case_study} presents two case studies from the Tokyo dataset—two traffic accidents and two traffic regulations—comparing ConFormer against AGGCN, ST-GCN, STAEFormer, and ground truth. In both accident cases (\figref{app:case1} and \figref{app:case2}), ConFormer accurately captured the traffic velocity drops at 12:00 and 09:00 respectively, significantly outperforming competing models in predicting these sudden deceleration patterns.

\begin{figure}[t]
	\centering
	\includegraphics[width=1\linewidth]{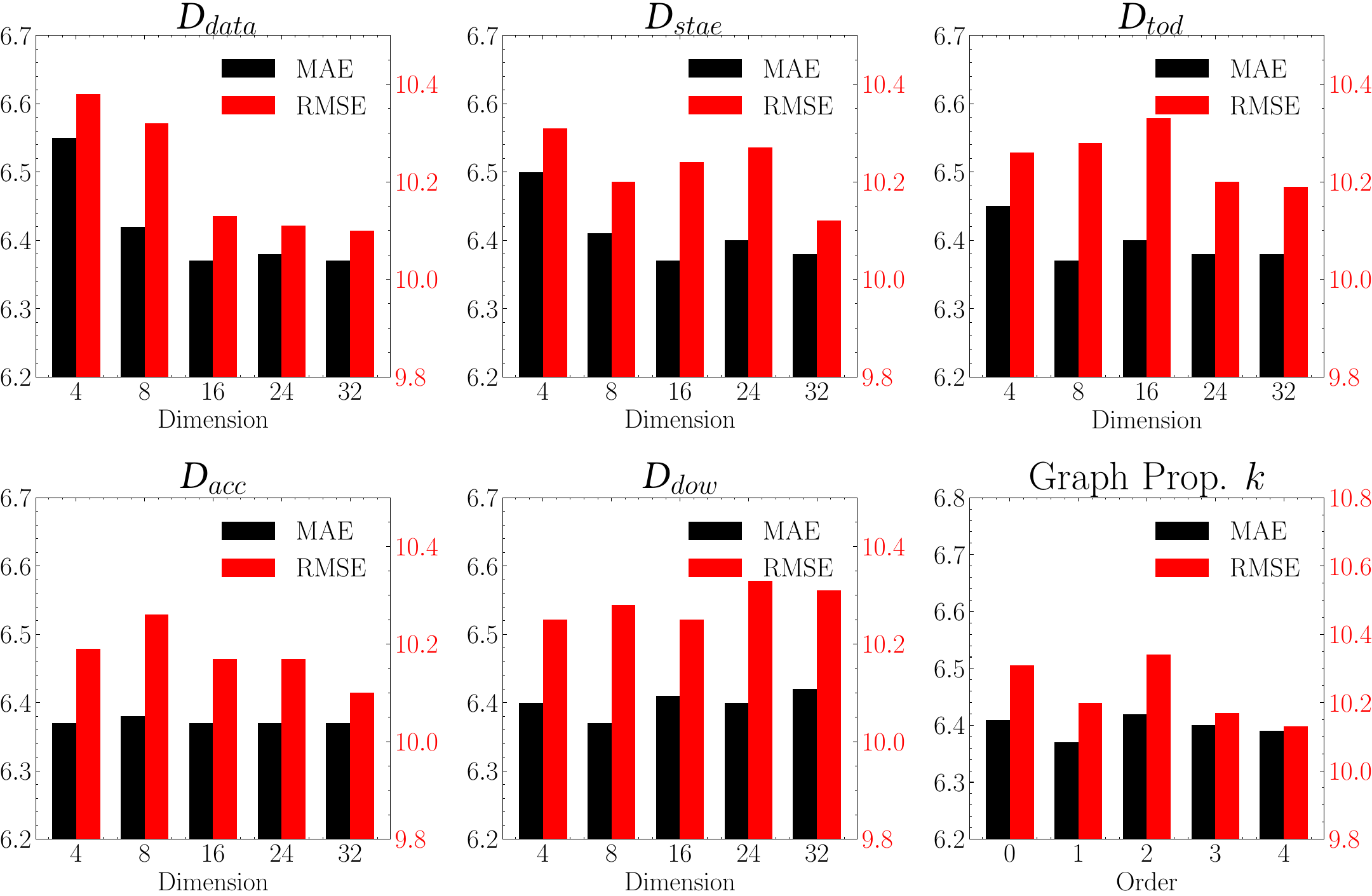}
	\caption{Hyperparameter study on Tokyo dataset.}
	\label{fig:hyperparameter}
\end{figure}

\subsection{Hyperparameter Study}  
We evaluated the sensitivity of the ConFormer architecture to key hyperparameters using the Tokyo dataset (\figref{fig:hyperparameter}). The analysis covered input embedding dimension ($D_{data}$), spatiotemporal attention dimension ($D_{stae}$), daily and weekly pattern dimensions ($D_d$, $D_w$), accident and regulation dimensions ($D_{acc}$, $D_{reg}$), and graph propagation order ($k$).  
Results show optimal performance with embedding dimensions between 8 and 16, with $D_{data}$ and $D_{stae}$ being most critical, while $D_{acc}$ and $D_{reg}$ had minimal impact. Shallow graph propagation (order 1–2) effectively captured spatial dependencies, with limited benefit from higher orders.  
These findings highlight ConFormer’s robustness and efficiency, ensuring adaptability to real-world scenarios with constrained resources.  

\section{Conclusion}
This paper introduces ConFormer, a novel conditional Transformer architecture for traffic prediction. By incorporating graph-based propagation adaptive normalization layers, ConFormer dynamically adjusts spatial and temporal correlations based on historical conditions. 
We validate our approach using two new datasets from Tokyo and California, demonstrating consistent superiority over mainstream spatio-temporal baselines.
ConFormer exhibits exceptional performance on both traditional PEMS benchmarks and our newly collected large-scale datasets with traffic incident information.




\bibliographystyle{ACM-Reference-Format}
\balance
\bibliography{reference}

\appendix
\setcounter{figure}{0}    
\setcounter{table}{0}
\setcounter{equation}{0}
\section{Appendix}

\subsection{Dataset Details}\label{sec:detail}

We use this traffic accident dataset \cite{moosavi2019countrywide} covering 49 states in the United States to match California's traffic network in Caltrans Performance Measurement System (PEMS)  with traffic accident data. This dataset has been collecting data from multiple data providers since February 2016. These data providers include multiple APIs that provide real-time traffic incident data streams. The traffic incidents broadcast by these APIs are captured by various entities, such as US and state transportation departments, law enforcement agencies, traffic cameras, and traffic sensors within the road network. Currently, the dataset contains about 1.5 million accident records.

Compared with PEMS dataset,  Tokyo provides traffic regulations and accidents, which includes fields for various types of lane restrictions, such as single lane restriction, double lane restriction, no restriction, under restriction, general lane restriction, and road closure. The dataset also records the causes of accidents, including overturned vehicle accidents, broken-down vehicles, accidents, vehicle fires, collision accidents, contact accidents, and rear-end collisions. An regulations dataset, includes additional causes such as cargo collapse, unknown causes, construction, scattered objects, earthquakes, falling objects, and road obstructions, along with a new regulation category for traffic jams. This dataset is instrumental for analyzing traffic accidents, understanding road conditions, and enhancing traffic management and safety measures.




\subsection{Training Details}
Our experiments ran on a GPU server with eight GeForce GTX 3090 graphics cards, employing the PyTorch 2.0.3 framework. The raw data was standardized using z-score normalization \cite{cheadle2003analysis}. Training was halted prematurely if validation error stabilized within 15-20 epochs or did not improve after 200 epochs, preserving the best model based on validation data \cite{luo2023dynamic}. We adhered to the original paper's model parameters and settings, while also conducting multiple parameter tuning iterations to enhance experimental outcomes. Data were partitioned chronologically into training, validation, and test sets at a 6:2:2 ratio across all sub-datasets. Model performance was assessed using Mask-Based Root Mean Square Error (RMSE), Mean Absolute Error (MAE), and Mean Absolute Percentage Error (MAPE) as metrics, disregarding zero values which indicated noisy data \cite{MegaCRN}. Recent research \cite{LargeST} indicates that existing Graph Convolutional Networks (GCNs) and Transformer-based methods are largely inadequate for real-world road networks, particularly for large graphs exceeding 2000 nodes, due to the prohibitive computational costs associated with capturing both global and local traffic patterns. As illustrated in \figref{tab:performance}, the majority of methods encounter out-of-memory errors when applied to the Bay Area dataset, which spans a one-year period. To address this challenge, we have implemented efficient linear attention mechanisms \cite{katharopoulos2020transformers,wang2020linformer,wu2024simplifying} on the Bay Area dataset. These mechanisms are designed to mitigate the substantial resource demands of traditional dot-product attention, which exhibits quadratic memory and computational complexity. The high computational and memory costs render dot-product attention impractical for real-world traffic networks. Our findings demonstrate that the utilization of linear attention does not compromise model performance.

\subsection{Baseline Details}
We included the following representative baselines in our study:
Historical Last (HL) \cite{liang2021revisiting}: A naive method that uses the average of historical values as the prediction result.
LSTM \cite{LSTM}: Focuses solely on temporal aspects, ignoring spatial correlations.
For methods combining GNNs \cite{defferrard2016convolutional, kipf2017semi}, we considered:
RNN-based approaches such as DCRNN \cite{DCRNN} and AGCRN \cite{AGCRN}.
TCN-based methods like STGCN \cite{stgcn}  and GWNET \cite{GWNet}.
Attention-based models including ASTGCN \cite{ASTGCN} and STTN \cite{xu2020spatial}.
Transformer variants such as PDFormer \cite{PDFormer} and STAEFormer \cite{STAEformer}.
Additionally, we integrated more sophisticated techniques:
Controlled differential-based method: STGODE \cite{STGODE} uses neural ordinary differential equations to model continuous traffic signal changes.
Dynamic graph-based methods: These include DSTAGNN \cite{DSTAGNN}, DGCRN \cite{li2023dynamic}, and D$^2$STGNN \cite{D2STGNN}, which specifically address dynamic correlations among sensors in traffic networks.



\subsection{Proof of Reformulated Self-Attention}\label{sec:thom}
With GLN, the vanilla self-attention can be reformulated as follows:
\begin{equation}\label{eq:appdenidx-1}
	\adjustbox{max width=\linewidth}{
		$\operatorname{Softmax}\left( \frac{\mathbf{Q} \mathbf{K}^\top}{\sqrt{D_k}} \right) \rightarrow \operatorname{Softmax}\left( \frac{\gamma^2 \mathbf{Q}' \mathbf{K}'^\top + \gamma \mathbf{K}' \beta^\top + \gamma \mathbf{Q}' \beta^\top + \beta^2}{\sqrt{D_k}} \right)$
	}
\end{equation}
where \(\mathbf{Q}' = \gamma \cdot \frac{\mathbf{Q} - \boldsymbol{\mu}_{\mathbf{Q}}}{\boldsymbol{\sigma}} + \beta\) and \(\mathbf{K}' = \gamma \cdot \frac{\mathbf{K} - \boldsymbol{\mu}_{\mathbf{K}}}{\boldsymbol{\sigma}} + \beta\).

\begin{proof}
	We proceed through several key steps to establish this theorem, leveraging the properties of Generalized Layer Normalization (GLN) and the linearity of the embedding layer.
	
	\begin{itemize}[leftmargin=0.4cm]
		\item \textbf{Normalization of Input:} For an input $x$, the normalized version using GLN with conditional affine transformation parameters $\gamma$ and $\beta$ is given by:
		\[
		x_i' = \gamma \cdot \frac{x_i-\mu_i}{\sigma_i} + \beta,
		\]
		where $\mu_i$ and $\sigma_i$ represent the mean and standard deviation of $x_i$, respectively.
		
		\item \textbf{Linear Embedding Transformation:} Given the linearity of the embedding layer, denoted by function $f$, we can express the normalized input as:
		\[
		q_i' = \gamma \cdot \frac{f(x_i) - f(\mu_i)}{\sigma_i} + f(\beta),
		\]
		where we define $\mu_{q_i} = f(\mu_i)$ and $\sigma_{q_i} = \sigma_i$. Thus, $q_i'$ can be rewritten as:
		\[
		q_i' = \gamma \cdot \frac{q_i - \mu_{q_i}}{\sigma_i} + \beta, \quad \text{where } \beta = f(\beta).
		\]
		
		\item \textbf{Formulation of $\mathbf{Q}'$ and $\mathbf{K}'$:} \\ 
		\begin{equation*}
			\scalebox{0.98}{$ 	\mathbf{Q}' = \gamma \cdot \frac{\mathbf{Q} - \boldsymbol{\mu}_{\mathbf{Q}}}{\boldsymbol{\sigma}} + \beta, \quad \mathbf{K}' = \gamma \cdot \frac{\mathbf{K} - \boldsymbol{\mu}_{\mathbf{K}}}{\boldsymbol{\sigma}} + \beta.$}
		\end{equation*}
		
		\item \textbf{Computation of $\mathbf{Q}' \mathbf{K}'^\top$:} \\ 
		\begin{equation*}
			\scalebox{0.98}{
				$\mathbf{Q}' \mathbf{K}'^\top = \left(\gamma \cdot \frac{\mathbf{Q} - \boldsymbol{\mu}_{\mathbf{Q}}}{\boldsymbol{\sigma}} + \beta\right) \left(\gamma \cdot \frac{\mathbf{K} - \boldsymbol{\mu}_{\mathbf{K}}}{\boldsymbol{\sigma}} + \beta\right)^\top$}
		\end{equation*}

		\item \textbf{Expansion of $\mathbf{Q}' \mathbf{K}'^\top$:} \\ 
		\begin{equation*}
			\scalebox{0.98}{$\mathbf{Q}' \mathbf{K}'^\top = \gamma^2 \left(\frac{\mathbf{Q} - \boldsymbol{\mu}_{\mathbf{Q}}}{\boldsymbol{\sigma}} \right) \left(\frac{\mathbf{K} - \boldsymbol{\mu}_{\mathbf{K}}}{\boldsymbol{\sigma}} \right)^\top + \gamma \beta \left(\frac{\mathbf{K} - \boldsymbol{\mu}_{\mathbf{K}}}{\boldsymbol{\sigma}} \right)^\top + \gamma \left(\frac{\mathbf{Q} - \boldsymbol{\mu}_{\mathbf{Q}}}{\boldsymbol{\sigma}} \right) \beta^\top + \beta^2.$}
		\end{equation*}
		
		\item \textbf{Formulation of the Softmax Attention:} \\ 
		Substituting this expanded form into the Self-Attention formula, we arrive at:
		\begin{equation*}
			\scalebox{0.98}{$
				\operatorname{Softmax}\left( \frac{\mathbf{Q} \mathbf{K}^\top}{\sqrt{D_k}} \right) = \operatorname{Softmax}\left( \frac{\gamma^2 \mathbf{Q}' \mathbf{K}'^\top + \gamma \mathbf{K}' \beta^\top + \gamma \mathbf{Q}' \beta^\top + \beta^2}{\sqrt{D_k}} \right).$}
		\end{equation*}
	\end{itemize}
	Thus, we have established the equivalence stated in Eq.~\ref{eq:appdenidx-1}, demonstrating how the normalized attention mechanism relates to the original formulation through a series of algebraic transformations.
\end{proof}

\end{document}